\documentclass[preprint,12pt]{elsarticle}

 \usepackage{url} 
\usepackage{amssymb}
\usepackage{subcaption}
\usepackage{multirow}
\usepackage{wrapfig}
\usepackage{booktabs}    
\usepackage{multirow}    
\usepackage{graphicx}    
\usepackage{array}       
\usepackage[table]{xcolor}
\definecolor{best}{RGB}{198, 226, 255}   
\definecolor{second}{RGB}{235, 245, 255} 
\newcommand{\best}[1]{\cellcolor{best}{\textbf{#1}}}
\newcommand{\secondbest}[1]{\cellcolor{second}{\underline{#1}}}
\usepackage{amsmath}

\journal{Nuclear Physics B}

\begin{document}

\begin{frontmatter}



\title{MissHDD: Hybrid Deterministic Diffusion for Hetrogeneous Incomplete Data Imputation}


\author{Youran Zhou, Mohamed Reda Bouadjenek, Sunil Aryal} 

\affiliation{organization={Deakin University},
            addressline={Waurn Ponds}, 
            city={Geelong},
            postcode={3216}, 
            state={Victoria},
            country={Australia}}

\begin{abstract}
Incomplete data are common in real-world tabular applications, where numerical, categorical, and discrete attributes coexist within a single dataset. This heterogeneous structure presents 
significant challenges for existing diffusion-based imputation models, which typically assume 
a homogeneous feature space and rely on stochastic denoising trajectories. Such assumptions make it difficult to maintain conditional consistency, and they often lead to information collapse for categorical variables or instability when numerical variables require deterministic updates. These limitations indicate that a single diffusion process is insufficient for mixed-type tabular imputation.

We propose a hybrid deterministic diffusion framework that separates heterogeneous features into two complementary generative channels. A continuous DDIM-based channel provides efficient and stable deterministic denoising for numerical variables, while a discrete latent-path diffusion channel, inspired by loopholing-based discrete diffusion, models categorical and discrete features without leaving their valid sample manifolds. The two channels are trained under a unified 
conditional imputation objective, enabling coherent reconstruction of mixed-type incomplete data.

Extensive experiments on multiple real-world datasets show that the proposed framework achieves higher imputation accuracy, more stable sampling trajectories, and improved robustness across MCAR, MAR, and MNAR settings compared with existing diffusion-based and classical methods. 
These results demonstrate the importance of structure-aware diffusion processes for advancing deep learning approaches to incomplete tabular data.
\end{abstract}

\begin{graphicalabstract}
\end{graphicalabstract}

\begin{highlights}
\item Research highlight 1
\item Research highlight 2
\end{highlights}


\end{frontmatter}


\section{Introduction}

Missing data are ubiquitous in real-world applications such as healthcare~\cite{healthcare}, 
finance~\cite{financial}, recommendation systems~\cite{recommendation}, and sensor networks~\cite{sensor}. 
In tabular datasets, missing entries can degrade predictive performance and introduce bias or 
uncertainty in downstream analysis. As a result, imputation, i.e., estimating missing values from 
partially observed data, remains a crucial step in modern data-centric pipelines. Classical 
approaches, including mean or mode substitution~\cite{mean,mean2}, 
$k$-nearest neighbors~\cite{knn}, MICE~\cite{mice}, and MissForest~\cite{missforest}, offer simple and 
computationally efficient solutions. However, they typically operate under restrictive assumptions, 
struggle to model complex feature dependencies, and adapt poorly to structured or mechanism-driven missingness.

To overcome the limitations of classical imputers, recent research has increasingly explored deep generative models for missing-data reconstruction. GAN-based methods such as
GAIN~\cite{yoon2018gain} and MisGAN~\cite{misGAN}, and variational approaches such as MIWAE~\cite{mattei2019miwae}, HI-VAE~\cite{hivae}, and their extensions~\cite{grape,igrm},
learn expressive joint distributions over observed and missing variables and have demonstrated notable improvements in imputation fidelity. In parallel, diffusion models have emerged as a powerful class of generative models with stable likelihood-based training and strong sample quality across vision, audio, and time-series domains~\cite{dm1,dm2,dreview1}. Several recent
works have adapted diffusion to the tabular setting, including TabCSDI~\cite{tabcsdi},
MissDiff~\cite{missdiff}, TabDDPM~\cite{tabddpm}, and DiffPuter~\cite{diffputer}. Denoising
Diffusion Implicit Models (DDIMs)~\cite{ddim} further provide a deterministic non-Markovian alternative to DDPMs, replacing stochastic transitions with parameterized reverse trajectories
that substantially reduce inference cost and sampling variance. These developments highlight the promise of diffusion-based methods for fine-grained tabular imputation through iterative
denoising.

Despite this progress, applying diffusion models to tabular imputation remains nontrivial~\cite{MLDL,Deeplearningisnotallyouneed}. Most existing methods build on DDPMs, whose stochastic sampling introduces high inference latency and produces different imputations across runs an undesirable property in applications requiring stability, determinism, or auditability. This is unsurprising: standard diffusion models were originally developed for generative tasks such as image or audio synthesis, where sampling diversity is essential to reflect the richness of the
data distribution. However, the objectives of tabular imputation differ fundamentally. Practitioners typically seek accurate and semantically consistent completions, or a reliable
surrogate dataset for downstream prediction or risk modeling. In this setting, excess sampling diversity provides little value and can instead harm reproducibility and downstream task
performance. Moreover, many diffusion-based imputers assume fully observed training samples or employ weak conditioning mechanisms that degrade when the inputs themselves are incomplete which is an
assumption violated by most real-world datasets.

A deeper challenge arises from the heterogeneous nature of tabular data. Conventional diffusion formulations are tailored to homogeneous continuous domains and do not naturally accommodate the mixed statistical manifolds of tabular features. Numerical, categorical, and discrete variables exhibit fundamentally different geometric and probabilistic structures. Applying a single continuous Gaussian diffusion process across all feature types leads to severe mismatches:
categorical variables suffer from semantic drift and information collapse, discrete states leave their valid support, and the model struggles to preserve conditional consistency across timesteps.
These structural incompatibilities form a central barrier to extending diffusion models to robust, heterogeneous tabular imputation.

In real-world settings, continuous features (e.g., laboratory measurements or financial indicators) commonly coexist with categorical or discrete attributes (e.g., diagnosis codes,
product categories, user identifiers). Numerical columns benefit from smooth deterministic denoising trajectories, whereas categorical and discrete variables must evolve on probability
simplexes to avoid invalid intermediate states. A single unified diffusion process is not suitable, as it must simultaneously satisfy incompatible geometric and probabilistic
requirements across feature types.

To address these challenges, we propose \textbf{MissHDD}, a hybrid deterministic diffusion framework
for imputing heterogeneous incomplete tabular data. MissHDD decomposes the generative process
into two complementary diffusion channels that respect the distinct statistical structure of
numerical and categorical/discrete features. A continuous DDIM-based channel handles numerical
variables through deterministic non-Markovian trajectories, enabling fast, stable, and
low-variance reconstruction. In parallel, a discrete latent-path diffusion channel—motivated by
recent loopholing-based discrete diffusion methods—operates on categorical and discrete variables
directly within their proper sample spaces, avoiding one-hot collapse and preserving
distributional structure across timesteps. These two channels are integrated under a unified
conditional objective that explicitly models the conditional distribution
\(p(x_{\mathrm{mis}} \mid x_{\mathrm{obs}})\), ensuring coherent conditioning throughout the
denoising process.

To support learning from realistically incomplete datasets, we further introduce a
\textbf{self-masking} strategy that creates pseudo-missing entries from partially observed samples.
This exposes the model to diverse missingness patterns and allows both diffusion channels to be
trained without requiring fully observed data—a critical property for real-world tabular domains
where complete samples are rare. Extensive experiments across multiple real-world datasets show
that MissHDD delivers state-of-the-art imputation accuracy, significantly improved run-to-run
stability, and substantially reduced inference latency compared with existing diffusion-based and
classical generative baselines.

\textbf{Our main contributions are as follows:}
\begin{itemize}
    \item We propose \textbf{MissHDD}, the first hybrid two-channel deterministic diffusion framework for heterogeneous incomplete tabular data, integrating a continuous DDIM channel for numerical features with a discrete latent-path diffusion channel for categorical and discrete features.
    
    \item We introduce a two-channel generative formulation that assigns feature types to diffusion processes aligned with their underlying statistical manifolds, ensuring stable conditioning and preventing information loss or semantic drift in categorical and discrete dimensions.
    
    \item We develop a unified self-masking training strategy that learns directly from incomplete data, eliminating the need for fully observed training samples and improving robustness to realistic MCAR, MAR, and MNAR missingness mechanisms.
    
    \item Through comprehensive evaluations on diverse real-world datasets, we demonstrate that MissHDD achieves state-of-the-art imputation accuracy while providing substantially faster, deterministic, and more reproducible inference than existing diffusion-based approaches.
\end{itemize}
\section{Related Work}

\textbf{Traditional Imputation Methods.}
Before the emergence of deep generative models, missing data imputation primarily relied on statistical heuristics and classical machine learning techniques. Common approaches include mean or mode substitution~\cite{mean,mean2}, $k$-nearest neighbors (KNN)~\cite{knn}, Multiple Imputation by Chained Equations (MICE)~\cite{mice}, and MissForest~\cite{missforest}. While computationally efficient and easy to deploy, these approaches often assume simple underlying data distributions and struggle to model complex feature correlations, especially under non-random missingness mechanisms. These limitations motivate the shift toward more expressive generative modeling approaches.

\textbf{Deep Generative Imputation.}
Generative Adversarial Networks (GANs) and Variational Autoencoders (VAEs) have been widely adopted for missing data imputation. Notable examples include GAIN~\cite{yoon2018gain} and MisGAN~\cite{misGAN}, which leverage adversarial training to match distributions of observed and imputed data. Variational inference–based models such as MIWAE~\cite{mattei2019miwae} and HI-VAE~\cite{hivae} extend these ideas by learning latent representations that capture joint dependencies across features. More recent architectures incorporate iterative refinement or graph-based inductive biases to improve expressiveness and relational modeling, such as GRAPE~\cite{grape} and IGRM~\cite{igrm}. Despite these advances, many generative models assume access to complete training data or rely on auxiliary networks tailored to specific missingness patterns. Furthermore, categorical and discrete and continuous variables are often treated jointly, without addressing their fundamentally different statistical structures.

\textbf{Diffusion Models for Missing Data.}
Diffusion models~\cite{dm1,dm2,dreview1} have achieved state-of-the-art performance in image synthesis, video generation, and time-series modeling, owing to their stable likelihood-based training and strong generative fidelity. Their success has motivated several adaptations for missing data. TabCSDI~\cite{tabcsdi} employs conditional score-based diffusion to impute masked entries, while MissDiff~\cite{missdiff} learns a data density model directly from partially observed data. DiffPuter~\cite{diffputer} integrates diffusion models into an EM-style framework to support conditional sampling. TabDDPM~\cite{tabddpm} extends diffusion modeling to mixed-type tabular data, but assumes fully observed training features, limiting its applicability.

A key limitation shared by most existing methods is their reliance on stochastic DDPM sampling, which requires hundreds of denoising steps and results in slow inference and variability across imputations. Recent deterministic formulations, such as our prior work MissDDIM~\cite{missddim}, alleviate this issue by replacing the stochastic reverse transitions with a non-Markovian DDIM-based sampler~\cite{ddim}, yielding faster and more consistent imputations. However, these methods still operate under a \emph{single} diffusion process for all feature types. This unified treatment overlooks the heterogeneous structure of tabular data, where numerical and categorical and discrete variables lie on fundamentally different manifolds. As a result, categorical and discrete features tend to suffer from probability–mass collapse or unstable transitions, and conditioning on observed values becomes inconsistent during the denoising trajectory.

\textbf{Generative Modeling for Heterogeneous Tabular Data.}
Tabular data typically contain both numerical and categorical and discrete attributes, each governed by distinct probability manifolds. This heterogeneity substantially complicates generative modeling. Several models aim to synthesize mixed-type tabular data using GANs, VAEs, or score-based methods~\cite{tvae,ctgan,tablegan}. More recent diffusion-based synthesizers address mixed-type domains using Gaussian transitions for continuous variables and multinomial transitions for discrete ones~\cite{tabddpm}. Discrete score matching methods~\cite{sun2022score,campbell2022continuous} also provide alternatives for categorical and discrete modeling. Nonetheless, existing approaches typically use a single unified generative process, which can be suboptimal for heterogeneous data due to representational mismatches between numerical and categorical and discrete attributes. This motivates hybrid architectures that treat different variable types through distinct generative pathways.

Our proposed \textbf{MissHDD} differs fundamentally from prior work in two important ways. First, unlike diffusion-based imputers that rely on stochastic DDPM sampling, MissHDD employs deterministic DDIM-based updates for numerical variables, offering stable and low-latency imputation. Second, instead of forcing heterogeneous features into a unified diffusion process, MissHDD introduces a discrete latent-path diffusion mechanism inspired by loopholing techniques to model categorical and discrete variables separately, preventing information collapse and improving conditional coherence. This two-channel design enables principled modeling of the conditional distribution \(p(x_{\text{mis}} \mid x_{\text{obs}})\) for mixed-type data. Finally, our self-masking training strategy allows the model to learn directly from incomplete data without requiring fully observed samples, addressing a limitation common to many diffusion-based approaches.

Collectively, MissHDD bridges the gap between expressive score-based generative modeling and the practical demands of heterogeneous incomplete tabular data imputation, offering improvements in accuracy, robustness, and inference efficiency over prior diffusion-based and classical methods.

\section{Preliminaries}

In this work, we focus on incomplete tabular datasets. Let
$\mathbf{X} \in \mathbb{R}^{n \times d}$ denote a dataset with $n$ samples
and $d$ features, where each row $\mathbf{x} \in \mathbb{R}^d$ may contain
missing values.

\subsection{Heterogeneous Feature Types in Tabular Data}

Tabular datasets often contain a mixture of continuous and categorical or
other discrete variables, each defined on distinct sample spaces and
governed by different statistical structures. This heterogeneity is central
to generative modeling because continuous and discrete features lie on
fundamentally different manifolds and therefore require different noise
processes, parameterizations, and sampling mechanisms.

Let $\mathbf{x} = (x_1, \ldots, x_d)$ denote a $d$-dimensional feature
vector. We partition its indices into two disjoint sets:
\[
\mathcal{I}_{\mathrm{cont}}, \quad \mathcal{I}_{\mathrm{cat}}, \qquad
\mathcal{I}_{\mathrm{cont}} \cup \mathcal{I}_{\mathrm{cat}} = \{1,\ldots,d\}, \;
\mathcal{I}_{\mathrm{cont}} \cap \mathcal{I}_{\mathrm{cat}} = \varnothing.
\]
For $i \in \mathcal{I}_{\mathrm{cont}}$, the feature is continuous,
\[
x_i \in \mathbb{R} \quad \text{or} \quad x_i \in \mathbb{R}^+,
\]
whereas for $j \in \mathcal{I}_{\mathrm{cat}}$, the feature is
categorical/discrete,
\[
x_j \in \{1, \ldots, K_j\},
\]
with $K_j$ denoting the number of categories for feature $j$.

We denote the decomposed feature subsets by
\[
\mathbf{x}^{\mathrm{cont}}
= \{x_i : i \in \mathcal{I}_{\mathrm{cont}}\}, \qquad
\mathbf{x}^{\mathrm{cat}}
= \{x_j : j \in \mathcal{I}_{\mathrm{cat}}\}.
\]

This decomposition is essential for our method because continuous and discrete variables obey fundamentally different noise geometries. Continuous features can be naturally modeled using Gaussian forward processes and DDIM-style deterministic denoising trajectories. In contrast, categorical and other discrete features require transition kernels defined on finite state spaces or probability simplexes to ensure that sampling remains within their valid domains. This structural heterogeneity motivates the two-channel formulation in MissHDD: a continuous DDIM pathway for $\mathbf{x}^{\mathrm{cont}}$ and a discrete latent-path diffusion mechanism for $\mathbf{x}^{\mathrm{cat}}$, each tailored to the statistical manifold on which the corresponding variables reside.

\subsection{Training with Incomplete Tabular Data}

We represent the missingness pattern in $\mathbf{X}$ by a binary mask
matrix $\mathbf{M} \in \{0,1\}^{n \times d}$:
\[
M_{ij} =
\begin{cases}
0, & \text{if } X_{ij} \text{ is observed}, \\
1, & \text{if } X_{ij} \text{ is missing}.
\end{cases}
\]

For a given sample $\mathbf{x}$ with corresponding mask $\mathbf{m}$, we
define the observed and missing subsets as
\[
\mathbf{x}^{\mathrm{obs}} = \mathbf{x} \odot (1 - \mathbf{m}), \qquad
\mathbf{x}^{\mathrm{mis}} = \mathbf{x} \odot \mathbf{m},
\]
where $\odot$ denotes element-wise multiplication. In particular,
$\mathbf{x}^{\mathrm{obs}}$ and $\mathbf{x}^{\mathrm{mis}}$ may each contain
both continuous and categorical/discrete coordinates, depending on which
entries are missing. The missingness process is governed by latent factors $\Psi$ and is
described by the conditional distribution $
p(\mathbf{M}\mid \mathbf{X}, \Psi).
$
Depending on how $\mathbf{M}$ depends on the observed and unobserved
entries of $\mathbf{X}$, the missingness mechanism is typically categorized as follows~\cite{MNAR}.
\paragraph{Missing Completely at Random (MCAR)}
Missingness is independent of both the observed and unobserved values:
\[
p(\mathbf{M}\mid \Psi).
\]
A common example is when laboratory results are missing because a blood sample was misplaced during transport or a machine malfunction occurred. The absence of the measurement is unrelated to any patient attributes or the underlying test value.

\paragraph{Missing At Random (MAR)}
Missingness depends only on the observed part of the data:
\[
p(\mathbf{M}\mid \mathbf{X}^{\mathrm{obs}},\Psi),
\qquad \text{independent of }\mathbf{X}^{\mathrm{mis}}.
\]
For instance, patients with higher recorded income or more frequent prior visits may be more likely to undergo optional screening tests. In this case, the missingness pattern depends on observed socioeconomic or behavioral variables, but not on the unobserved measurement itself.

\paragraph{Missing Not At Random (MNAR)}
Missingness depends directly on the missing values:
\[
p(\mathbf{M}\mid \mathbf{X}^{\mathrm{mis}},\Psi),
\qquad \text{independent of }\mathbf{X}^{\mathrm{obs}}.
\]
A concrete example is when patients with abnormally high blood pressure or glucose levels tend not to return for follow-up visits, leading to missing clinical measurements precisely because the unobserved values are extreme or unfavourable.
\medskip

Given an incomplete sample $\mathbf{x}_0$ with decomposition
$(\mathbf{x}_0^{\mathrm{obs}}, \mathbf{x}_0^{\mathrm{mis}})$, the goal of
imputation is to model the conditional distribution:
$
p(\mathbf{x}^{\mathrm{mis}}_0 \mid \mathbf{x}^{\mathrm{obs}}_0)
$
and infer plausible values for the missing entries using the observed features. 

\subsection{Denoising Diffusion Probabilistic Models}
\label{sec:background:ddpm}

Denoising Diffusion Probabilistic Models (DDPMs)~\citep{ddpm} are a class
of latent-variable generative models that learn data distributions by
gradually destroying structure through a forward noising process and then
learning to invert this corruption via a reverse denoising process. Let
$\mathbf{x}_0 \in \mathbb{R}^d$ denote a sample drawn from the unknown data
distribution $q(\mathbf{x}_0)$ (for example, a continuous subset of a
tabular sample). DDPMs construct a Markov chain of latent variables
\[
\mathbf{x}_0 \rightarrow \mathbf{x}_1 \rightarrow \cdots \rightarrow \mathbf{x}_T,
\]
where $T$ is a predefined number of diffusion steps.

\subsubsection{Forward diffusion}
In the forward process, small amounts of Gaussian noise are incrementally
added:
$
q(\mathbf{x}_t \mid \mathbf{x}_{t-1})
= \mathcal{N}\!\left(
\mathbf{x}_t;\sqrt{1-\beta_t}\,\mathbf{x}_{t-1},\,\beta_t \mathbf{I}
\right),
$
where $\{\beta_t\}_{t=1}^T$ is a variance schedule that controls the rate
at which information is destroyed. By composing these transitions, the
forward process admits the closed-form marginal
\begin{equation}
q(\mathbf{x}_t \mid \mathbf{x}_0)
= \mathcal{N}\!\left(
\mathbf{x}_t;\sqrt{\alpha_t}\,\mathbf{x}_0,\,(1-\alpha_t)\mathbf{I}
\right),
\qquad
\alpha_t = \prod_{i=1}^t (1-\beta_i),
\label{eq:ddpm-marginal}
\end{equation}
so $\mathbf{x}_t$ can be sampled directly from $\mathbf{x}_0$ without
simulating all intermediate steps. As $t$ approaches $T$, the distribution
of $\mathbf{x}_t$ converges to an isotropic Gaussian, effectively removing
all information about $\mathbf{x}_0$.

\subsubsection{Reverse denoising process}
The generative model learns to reverse this corruption by parameterizing
the reverse transitions as
\begin{equation}
p_\theta(\mathbf{x}_{t-1} \mid \mathbf{x}_t)
= \mathcal{N}\!\left(
\mathbf{x}_{t-1};
\,\boldsymbol{\mu}_\theta(\mathbf{x}_t, t),\,
\sigma_\theta^2(t)\mathbf{I}
\right),
\label{eq:ddpm-reverse}
\end{equation}
where the mean is computed from a neural network that predicts the
injected noise:
\begin{equation}
\boldsymbol{\mu}_\theta(\mathbf{x}_t, t)
= \frac{1}{\sqrt{\alpha_t}}
\left(
\mathbf{x}_t - \sqrt{1-\alpha_t}\,\epsilon_\theta(\mathbf{x}_t, t)
\right).
\label{eq:ddpm-mu}
\end{equation}
Thus, learning the generative model reduces to learning the noise
prediction function $\epsilon_\theta$, often referred to as the score
network. Sampling from the model requires iteratively applying the reverse
transitions from $t = T$ back to $t = 0$.

\subsubsection{Training objective}
Under a variational bound on the data likelihood, DDPMs can be trained
using the simplified denoising score-matching loss:
\begin{equation}
\mathcal{L}_{\text{DDPM}}(\theta)
= \mathbb{E}_{\mathbf{x}_0,t,\boldsymbol{\epsilon}}
\left[
\left\|
\boldsymbol{\epsilon}
- \epsilon_\theta\big(\sqrt{\alpha_t}\mathbf{x}_0
+ \sqrt{1-\alpha_t}\,\boldsymbol{\epsilon},\, t\big)
\right\|_2^2
\right],
\label{eq:ddpm-loss}
\end{equation}
where $\boldsymbol{\epsilon} \sim \mathcal{N}(\mathbf{0},\mathbf{I})$ and
$t$ is sampled from a predefined distribution (typically uniform on
$\{1,\ldots,T\}$). This objective trains the model to recover the noise
that produced $\mathbf{x}_t$ from $\mathbf{x}_0$.

\subsubsection{Limitations for imputation}
While DDPMs achieve excellent generative performance for continuous data,
the sampling process is inherently sequential and stochastic. A single
generation requires hundreds of reverse diffusion steps, leading to high
inference latency, and the stochasticity introduces variability across
runs. For imputation tasks on tabular data, where predictions must be fast,
stable, and reproducible. This motivates
the use of deterministic diffusion formulations such as DDIM~\citep{ddim,missddim}
and, their extension to conditional and heterogeneous imputation, as described in the next section.

\begin{figure*}[ht]
    \centering
\includegraphics[width=\linewidth, height=0.30\textheight, keepaspectratio]{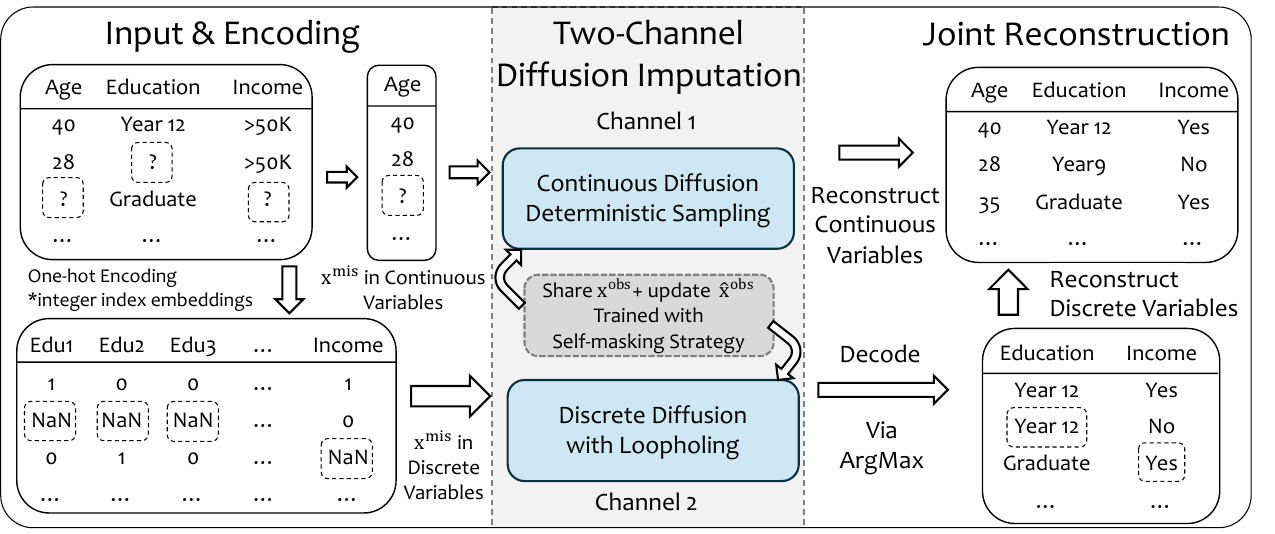}
\caption{
Overview of the \textbf{MissHDD} framework. 
Continuous and categorical and discrete features are separated and imputed through two diffusion channels: 
a conditional deterministic DDIM for continuous variables (Channel~1) and a Loopholing-based discrete diffusion model for categorical and discrete variables (Channel~2). 
categorical and discrete inputs are represented using one-hot encoding, while integer index embeddings are also supported. 
Both channels are trained with a self-masking strategy, and the reconstructed outputs are decoded and merged to form the final imputed table.
}
    \label{fig:MissHDD_overview}
\end{figure*}
\section{The Proposed MissHDD Framework}

To address the inefficiency, instability, and heterogeneous feature incompatibilities observed in
existing diffusion-based imputers, we introduce \textbf{MissHDD}, a \textbf{H}ybrid \textbf{D}eterministic \textbf{D}iffusion framework tailored for incomplete tabular data. MissHDD is built on the key insight that numerical
and categorical/discrete variables occupy fundamentally different statistical manifolds and therefore
cannot be reliably modeled under a single continuous diffusion process. Accordingly, MissHDD adopts
a two-channel architecture in which each feature type is modeled by a diffusion mechanism that matches
its underlying structure. The \emph{continuous channel} employs a conditional deterministic DDIM sampler that denoises only the
missing numerical coordinates while conditioning on the observed ones. This channel produces fast,
stable, and variance-free reconstructions by leveraging non-Markovian deterministic trajectories,
making it well suited for downstream pipelines that require reproducible imputations. The \emph{discrete channel} operates in parallel via a loopholing-based discrete diffusion mechanism
designed for categorical or other finite-state variables. Unlike continuous Gaussian diffusion, it
pushes one-hot vectors off their probability simplex and leads to semantic drift—the discrete channel preserves probability mass across timesteps and prevents the collapse behaviors commonly observed in naïve one-hot diffusion. This ensures that categorical and discrete variables remain within their
valid state space during both training and sampling. Both channels are unified under a shared conditional objective that models
\(
p(\mathbf{x}^{\mathrm{mis}} \mid \mathbf{x}^{\mathrm{obs}})
\)
and enforces coherent conditioning throughout the denoising trajectory. To enable learning directly
from realistically incomplete datasets, MissHDD further incorporates a \emph{self-masking} strategy
that constructs pseudo-missing targets from partially observed samples, eliminating any requirement
for fully observed training data. As illustrated in Figure~\ref{fig:MissHDD_overview}, the outputs of the two channels are decoded and
merged to produce the final imputed table. The following sections describe each component of the
MissHDD framework in detail.

\subsection{Channel 1: Conditional DDIM for Efficient Continuous Variable Imputation}

Numerical features in tabular data lie on a continuous Euclidean manifold where Gaussian noise
and linear diffusion dynamics are well defined. This makes diffusion models a natural choice for
learning conditional distributions over missing numerical values. However, standard
Denoising Diffusion Probabilistic Models (DDPMs)~\cite{ddpm} employ a stochastic Markovian
reverse process that requires hundreds of iterative denoising steps and injects randomness at each
transition. While such stochasticity is beneficial for diverse generative sampling, it is fundamentally
misaligned with the requirements of imputation, where outputs should be more efficient and stable. These limitations motivated the deterministic formulation introduced in DDIM~\cite{ddim}, and further demonstrated in our prior MissDDIM framework~\cite{missddim}, where we showed that
deterministic non-Markovian trajectories provide accurate and low-variance imputations for continuous tabular data. In MissHDD, we build upon these insights by extending DDIM to a \emph{conditional} diffusion
mechanism that reconstructs missing numerical coordinates while conditioning every denoising step on the observed portion of the input.

\subsubsection{Forward diffusion on missing coordinates}
Let $\mathbf{x}_0 \in \mathbb{R}^d$ be a tabular sample and $\mathbf{m} \in \{0,1\}^d$ the
missingness mask. We decompose the sample into observed and missing subsets:
\[
\mathbf{x}_0^{\mathrm{obs}}=\mathbf{x}_0 \odot (1-\mathbf{m}), \qquad
\mathbf{x}_0^{\mathrm{mis}}=\mathbf{x}_0 \odot \mathbf{m}.
\]
In the continuous channel, we operate only on the numerical coordinates within the
missing subset. Let
\[
\mathbf{x}_0^{\mathrm{mis,cont}}
= \{ x_i : i \in \mathcal{I}_{\mathrm{mis}} \cap \mathcal{I}_{\mathrm{cont}} \}.
\]
For notational simplicity, we denote this subset simply by
$\mathbf{x}^{\mathrm{mis}}$. 
Forward diffusion is applied only to the missing numerical dimensions., leaving
observed features untouched. The corrupted latent state at step $t$ is:
\[
\mathbf{x}_t^{\mathrm{mis}} =
\sqrt{\alpha_t}\,\mathbf{x}_0^{\mathrm{mis}}
+\sqrt{1-\alpha_t}\,\boldsymbol{\epsilon},\qquad
\boldsymbol{\epsilon}\sim\mathcal{N}(\mathbf{0},\mathbf{I}),
\]
where $\alpha_t=\prod_{s=1}^t(1-\beta_s)$ follows the variance schedule. This design allows the
model to explicitly learn the conditional distribution
\(
p(\mathbf{x}_0^{\mathrm{mis}} \mid \mathbf{x}_0^{\mathrm{obs}}),
\)
rather than modeling the full joint distribution of heterogeneous tabular data.

\subsubsection{Conditional reverse process.}
To recover the missing components, this Channel employs a conditional noise estimator:
\[
\epsilon_\theta(\mathbf{x}_t^{\mathrm{mis}},t\mid\mathbf{x}_0^{\mathrm{obs}}),
\]
which receives both the corrupted missing coordinates and the clean observed features. Under
DDPM, the reverse transition is defined as a Gaussian:
\[
p_\theta(\mathbf{x}_{t-1}^{\mathrm{mis}}\mid\mathbf{x}_t^{\mathrm{mis}},\mathbf{x}_0^{\mathrm{obs}})
=
\mathcal{N}\!\left(
\mathbf{x}_{t-1}^{\mathrm{mis}};\,
\boldsymbol{\mu}_\theta(\mathbf{x}_t^{\mathrm{mis}},t\mid\mathbf{x}_0^{\mathrm{obs}}),\,
\sigma_\theta^2(t)\mathbf{I}
\right),
\]
with mean
\[
\boldsymbol{\mu}_\theta =
\frac{1}{\sqrt{\alpha_t}}
\left(
\mathbf{x}_t^{\mathrm{mis}}
-\frac{\beta_t}{\sqrt{1-\alpha_t}}\,
\epsilon_\theta(\mathbf{x}_t^{\mathrm{mis}},t\mid\mathbf{x}_0^{\mathrm{obs}})
\right).
\]
Because $\sigma_t>0$ under DDPM, each transition injects fresh Gaussian noise, producing
irreducible sampling variance and slow convergence, two characteristics that contradict the
requirements of tabular imputation.

\subsubsection{Deterministic DDIM trajectories}
DDIM introduces a family of non-Markovian reverse-time trajectories that remain \emph{consistent}
with the objective optimized during DDPM training. Formally, DDIM reinterprets the DDPM reverse
process as a discretization of an underlying generative ODE. Among the family of admissible
solutions, setting the stochasticity parameter $\sigma_t=0$ yields a deterministic trajectory:
\[
\mathbf{x}_{t-1}^{\mathrm{mis}} =
\sqrt{\alpha_{t-1}}
\left(
\frac{\mathbf{x}_t^{\mathrm{mis}}
      -\sqrt{1-\alpha_t}\,
       \epsilon_\theta(\mathbf{x}_t^{\mathrm{mis}},t\mid\mathbf{x}_0^{\mathrm{obs}})}
     {\sqrt{\alpha_t}}
\right)
+
\sqrt{1-\alpha_{t-1}}\,\epsilon_\theta(\mathbf{x}_t^{\mathrm{mis}},t\mid\mathbf{x}_0^{\mathrm{obs}}).
\]
This update eliminates all stochastic sampling noise while preserving the DDPM training objective,
because DDIM trajectories correspond to exact solutions of the reverse-time ODE implied by the
score model. A key consequence of adopting DDIM sampling is that the reverse mapping 
$\mathbf{x}_T^{\mathrm{mis}} \rightarrow \mathbf{x}_0^{\mathrm{mis}}$ becomes fully deterministic, 
yielding imputations that are stable and reproducible across runs. Moreover, DDIM supports large 
non-uniform timestep skips (often 20-50 steps)~\cite{missddim} without compromising reconstruction quality, reducing sampling cost by an order of magnitude compared with DDPM. Finally, the absence of stochastic noise in the reverse trajectory eliminates variance accumulation, producing smoother and 
more numerically stable denoising paths these properties that are particularly important for reliable 
downstream inference and decision-making tasks.

\subsubsection{Training objective.}
The continuous channel of MissHDD follows the standard DDPM/DDIM training paradigm, 
adapted to the conditional imputation setting. Given a sample with observed numerical 
features $\mathbf{x}_0^{\mathrm{obs}}$ and missing targets $\mathbf{x}_0^{\mathrm{mis}}$, only the missing 
coordinates are corrupted by the forward diffusion process:
\[
\mathbf{x}_t^{\mathrm{mis}}
= \sqrt{\alpha_t}\,\mathbf{x}_0^{\mathrm{mis}}
  + \sqrt{1-\alpha_t}\,\boldsymbol{\epsilon},
\qquad 
\boldsymbol{\epsilon}\sim\mathcal{N}(\mathbf{0},\mathbf{I}),
\]
while the observed entries remain fixed and are supplied as conditioning information 
throughout the reverse steps.

A conditional noise-prediction network
\[
\epsilon_\theta(\mathbf{x}_t^{\mathrm{mis}},\, t \mid \mathbf{x}_0^{\mathrm{obs}})
\]
is trained to recover the injected Gaussian noise. This leads to the conditional 
denoising score-matching objective:
\[
\min_{\theta}\;
\mathbb{E}_{\mathbf{x}_0,\, t,\, \boldsymbol{\epsilon}}
\left[
\left\|
\boldsymbol{\epsilon}
-
\epsilon_\theta(\mathbf{x}_t^{\mathrm{mis}},\, t \mid \mathbf{x}_0^{\mathrm{obs}})
\right\|_2^2
\right],
\]
where $t$ is sampled uniformly from $\{1,\dots,T\}$.  
This formulation trains the model to approximate the conditional score of the missing 
components, enabling the deterministic DDIM sampler to reconstruct 
$\mathbf{x}_0^{\mathrm{mis}}$ from $\mathbf{x}_T^{\mathrm{mis}}$ in a stable, efficient, and reproducible manner.

\subsubsection{Limitations for categorical and discrete variables}

Although DDIM offers an efficient and stable mechanism for continuous numerical features, its 
formulation is fundamentally misaligned with the nature of categorical and other discrete 
variables. Categorical values reside on a finite, unordered state space where Gaussian 
perturbations and linear interpolation have no semantic meaning. As soon as the DDIM forward 
step $\mathbf{x}_t=\sqrt{\alpha_t}\mathbf{x}_0+\sqrt{1-\alpha_t}\,\boldsymbol{\epsilon}$ is applied, 
a one-hot vector or index embedding is mapped to a dense point in $\mathbb{R}^K$ that no longer 
corresponds to any valid category, removing the sample from the discrete manifold at the very 
first timestep.

Because these intermediate noisy states are not legitimate categories, the reverse process becomes 
ill-defined: the model must project ambiguous continuous vectors back onto a discrete set without 
a meaningful geometric structure to guide the mapping. This often results in semantic drift, 
collapsing class boundaries, and unstable denoising dynamics—well-documented failure modes of 
continuous diffusion when applied to discrete domains. Moreover, DDIM relies on deterministic 
ODE-based trajectories, which inherently assume a continuous underlying space; such trajectories 
do not exist on discontinuous categorical manifolds without imposing continuous relaxations that 
weaken categorical fidelity.

These structural incompatibilities make continuous Gaussian diffusion unsuitable for discrete 
variables. Effective imputation in this setting requires a diffusion process defined directly on 
discrete probability simplexes, where transitions and denoising dynamics remain within the valid 
state space.

This motivates the second core component of MissHDM: a loopholing-based discrete diffusion 
channel designed to model categorical and discrete variables natively. In the next section, we 
describe how this discrete channel complements the continuous DDIM pathway to form a unified 
heterogeneous imputation framework.

\subsection{Channel 2: Conditional Loopholing Diffusion for Categorical and Discrete Variables}

While the continuous DDIM channel provides an efficient mechanism for imputing numerical
features, categorical and other discrete variables require a fundamentally different generative
process. As discussed earlier, Gaussian perturbations, linear interpolation, and continuous ODE
trajectories are incompatible with discrete state spaces: a single diffusion step already pushes a
categorical value off its valid manifold, making the reverse denoising process ill-posed.
Accordingly, the second component of MissHDM adopts a \emph{loopholing-based discrete diffusion}
framework~\cite{loopholing}, which was originally proposed for unconditional discrete generation.
Loopholing diffusion defines its forward and reverse transitions directly on discrete probability
simplexes, ensuring that all intermediate states remain valid categorical distributions and avoiding
the probability–mass collapse commonly observed in one-hot diffusion. In MissHDM, we extend this framework to a \emph{conditional} form tailored to tabular imputation: the reverse denoising updates are conditioned on the observed features, and the forward corruption is applied only to the missing categorical coordinates. This conditional adaptation aligns the loopholing mechanism with the imputation objective, enabling coherent conditioning across timesteps and producing stable, semantically consistent reconstructions for categorical and discrete variables.

Each sample from the discrete subset takes the form
\[
\mathbf{x}_0 \in \{1,\ldots,K_1\}\times\cdots\times\{1,\ldots,K_{d_{\mathrm{cat}}}\},
\] with a corresponding mask $\mathbf{m}$ indicating which discrete coordinates require imputation. For each variable $x_j$, the model internally supports two equivalent representations: a one-hot vector $\mathbf{e}_{x_j} \in \{0,1\}^{K_j}$, or an integer index passed through a learnable embedding matrix. Both formats are naturally integrated in the proposed framework because loopholing diffusion operates over probability simplexes rather than Euclidean coordinates, allowing categorical distributions to be manipulated directly without continuous relaxations. In the discrete channel, we operate only on the categorical/discrete coordinates within the
missing subset. Let
\[
\mathbf{x}_0^{\mathrm{mis,cat}}
= \{ x_i : i \in \mathcal{I}_{\mathrm{mis}} \cap \mathcal{I}_{\mathrm{cat}} \}.
\]
For notational simplicity, we denote this subset simply by
$\mathbf{x}^{\mathrm{mis}}$.

\subsubsection{Forward Process: Loopholing Corruption on categorical and discrete Simplexes.}
For each missing categorical and discrete coordinate, the forward diffusion step transforms the one-hot 
representation into a smoothed categorical and discrete distribution. Specifically, denote the categorical and discrete state 
at time $t$ as a simplex vector $\mathbf{p}_t^{(j)} \in \Delta^{K_j-1}$. The forward kernel is defined as
\begin{equation}
q(\mathbf{p}_{t}^{(j)} \mid \mathbf{p}_{t-1}^{(j)})
=(1-\beta_t)\mathbf{p}_{t-1}^{(j)}+\beta_t \mathbf{u}^{(j)},
\label{eq:forward_cat}
\end{equation}
where $\mathbf{u}^{(j)}$ is the uniform distribution over categories. This loopholing mechanism 
keeps $\mathbf{p}_t^{(j)}$ inside the simplex at all times, avoids invalid continuous vectors, and 
ensures that the transition preserves class boundaries. For observed categories, the state is kept 
fixed and supplied as conditional context.

\subsubsection{Conditional Reverse Process.}
Given corrupted categorical and discrete distributions $\mathbf{p}_t^{\mathrm{mis}}$, the reverse process predicts the 
clean categorical and discrete distribution at step $t-1$ using a conditional classifier
\[
f_\theta(\mathbf{p}_t^{\mathrm{mis}}, t \mid \mathbf{x}^{\mathrm{obs}}),
\]
which outputs logits corresponding to categorical and discrete transitions. The reverse kernel adopts the 
loopholing form:
\begin{equation}
p_\theta(\mathbf{p}_{t-1}^{(j)} \mid \mathbf{p}_t^{(j)}, \mathbf{x}^{\mathrm{obs}})
\propto
\exp\!\left(
f_\theta^{(j)}(\mathbf{p}_t^{(j)}, t \mid \mathbf{x}^{\mathrm{obs}})
\right)
\odot
\left((1-\beta_t)\mathbf{p}_t^{(j)}+\beta_t\mathbf{u}^{(j)} \right),
\label{eq:reverse_cat}
\end{equation}
where $\odot$ denotes elementwise multiplication. This structure ensures that the predicted reverse-time distribution remains entirely within the
categorical and discrete simplex, faithfully mirrors the form of the forward transition kernel, and
naturally incorporates observed features as conditioning signals. At the final step ($t=0$), the
model produces a valid discrete output by selecting the most probable category for each variable
via an $\arg\max$ operation.

Loopholing diffusion is well suited to categorical and discrete imputation because its transitions are defined directly on the probability simplex, ensuring that all intermediate states remain valid categorical distributions. This manifold preservation prevents the semantic drift typically observed
when continuous diffusion is applied to discrete variables, allowing class boundaries to remain well defined throughout the trajectory. Moreover, the reverse denoising steps can incorporate the observed features as conditioning signals without relying on continuous relaxations or surrogate
Gaussian noise. Importantly, the framework treats both one-hot and index-based embeddings consistently, as both representations are mapped to the same simplex space prior to diffusion, yielding a unified and structurally coherent approach to discrete imputation.

\subsubsection{Training Objective.}
The model is trained using a standard cross-entropy denoising objective:
\begin{equation}
\min_{\theta} 
\mathbb{E}_{\mathbf{x}_0, t}
\Big[
\mathrm{CE}\big(
\mathbf{p}_0^{\mathrm{mis}},
f_\theta(\mathbf{p}_t^{\mathrm{mis}}, t \mid \mathbf{x}^{\mathrm{obs}})
\big)
\Big],
\end{equation}
where $\mathbf{p}_0^{\mathrm{mis}}$ is the ground-truth one-hot distribution for missing categories.

Through this conditional loopholing diffusion mechanism, Channel 2 enables MissHDD to perform 
categorical and discrete imputation in a manner that preserves discrete structure, avoids invalid continuous 
states, and complements the efficient numerical imputation provided by the continuous DDIM channel.

\begin{figure}[ht]
    \centering
    \includegraphics[width=0.8\linewidth]{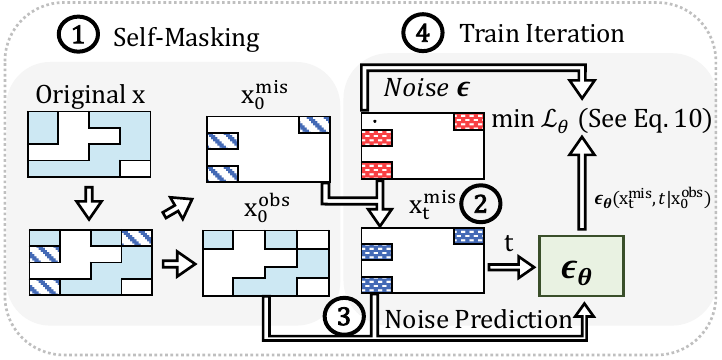}
    \caption{
    Self-masking strategy used in MissHDD.  
    During training, a portion of observed entries is randomly masked to create 
    pseudo-missing targets. The remaining entries serve as conditioning information for 
    both the continuous and discrete diffusion channels.
    }
    \label{fig:fig_self_masking}
\end{figure}
\subsection{Self-masking Strategy for Incomplete Data}

Learning from incomplete tabular data is challenging because only a subset of the true entries is observed, and relying on external imputers (e.g., mean filling or KNN) introduces bias before the generative model is trained. To avoid such preprocessing, MissHDM adopts a \textbf{self-masking strategy} that enables both diffusion channels to learn directly from partially observed samples. For each input, the original missingness pattern is preserved to respect the true underlying mechanism. In addition, a small random subset of the observed entries is temporarily masked during training to serve as pseudo-missing targets, while the remaining observed values are used as conditioning information. This simple procedure increases the amount of supervised signal available to the model and regularizes both diffusion channels by exposing them to slightly perturbed masking patterns, without altering the structure of the original missingness. As illustrated in Figure~\ref{fig:fig_self_masking},
the same self-masked sample is passed to both the continuous DDIM channel and the loopholing-based discrete channel, ensuring that the model consistently learns the conditional
distribution \(p(\mathbf{x}^{\mathrm{mis}}_0 \mid \mathbf{x}^{\mathrm{obs}}_0)\) under realistic training conditions.

\subsection{Joint Reconstruction}

After both diffusion channels complete their conditional denoising trajectories, the
numerical and categorical/discrete estimates must be combined into a single, coherent
imputed sample. In MissHDM, this is not performed through a naive concatenation.
Instead, the two channels are inherently coupled through a shared conditioning
mechanism: at every reverse diffusion step, each channel conditions on
(i) the observed features $\mathbf{x}^{\mathrm{obs}}$, and 
(ii) the latest estimates produced by the other channel. 
This shared-conditioning design ensures that heterogeneous feature types remain
statistically aligned throughout the denoising process.

At the final step $t=0$, the continuous channel outputs a deterministic reconstruction
$\hat{\mathbf{x}}^{\mathrm{cont}}$, while the discrete channel yields categorical probability
vectors from which the final discrete values $\hat{\mathbf{x}}^{\mathrm{cat}}$ are obtained via
\(\arg\max\). The full imputed sample is then formed as
\[
\hat{\mathbf{x}}
=
\hat{\mathbf{x}}^{\mathrm{cont}}
\,\cup\,
\hat{\mathbf{x}}^{\mathrm{cat}},
\]
with the union respecting the original feature ordering.

Because the diffusion trajectories of both channels are guided by the same conditioning
signals, the resulting reconstruction corresponds to a jointly consistent estimate of
$p(\mathbf{x}^{\mathrm{mis}} \mid \mathbf{x}^{\mathrm{obs}})$ rather than a pair of independent
imputations. This guarantees that cross-type dependencies—such as the relationship
between categorical attributes and numerical measurements—are preserved in the final
imputed dataset.

section{Experimental Setup}

We evaluate \texttt{MissHDM} along four complementary dimensions:
(i)\textbf{Imputation accuracy}, measuring how closely the imputed values recover the ground truth;
(ii)\textbf{Downstream task utility}, assessing whether the reconstructed data preserves predictive structure for supervised learning;
(iii)\textbf{Sampling efficiency}, quantifying the trade-off between inference cost and the number of denoising steps; and
(iv)\textbf{Stability}, examining the consistency of imputations across repeated runs.
Together, these criteria provide a comprehensive assessment of both the statistical fidelity and practical reliability of the proposed heterogeneous diffusion framework.


\subsubsection{Datasets}

We evaluate \texttt{MissDDIM} on sixteen heterogeneous tabular datasets drawn from the UCI Machine Learning Repository\footnote{\url{https://archive.ics.uci.edu/}} and Kaggle\footnote{\url{https://www.kaggle.com/}}.  
Table~\ref{tab:dataset_summary} summarizes their key characteristics.  
These datasets cover diverse domains, including healthcare, finance, materials, and sensor analytics and exhibit a wide range of sample sizes, dimensionalities, and proportions of categorical versus continuous features.  
Such heterogeneity makes them well suited for assessing diffusion-based imputers designed to operate on mixed-type tabular data.

For datasets accompanied by class labels, we evaluate both imputation quality and downstream classification performance.  
Unlabeled datasets are used exclusively for reconstruction accuracy.  
This dual setting allows a comprehensive assessment of \texttt{MissDDIM} as both an imputer and a 
preprocessing tool for predictive modeling.

\subsubsection{Missing Data Generation}

To benchmark \texttt{MissDDIM} under controlled yet realistic conditions, we introduce missingness
using the three canonical mechanisms \textbf{MCAR}, \textbf{MAR}, and \textbf{MNAR} at
missing rates of 30\%, 50\%, and 70\%.  
All masks are generated using the \texttt{MissMecha} package~\cite{missmecha}, which provides
reproducible and mechanism-consistent missingness simulation for heterogeneous tabular data. Using \texttt{MissMecha} ensures that the structural assumptions of each mechanism are applied
systematically and identically across all competing models, enabling fair comparison—particularly important for diffusion-based imputers that are sensitive to masking patterns.

\paragraph*{MCAR (Missing Completely at Random)}
Entries are removed uniformly across all samples and features, independent of both observed and 
unobserved values.  
This setting provides an unbiased baseline to examine how well models handle unstructured data loss.

\paragraph*{MAR (Missing at Random)}
The probability of removing a feature \(x_k\) depends on the value of an observed covariate 
\(x_{k'}\).  
Following~\cite{ot}, entries in \(x_k\) are masked when \(x_{k'}\) exceeds or falls below a 
dataset-specific threshold (e.g., mean or percentile).  
This mechanism reflects practical scenarios where missingness is driven by observable attributes.

\paragraph*{MNAR (Missing Not at Random)}
Missingness depends on the unobserved value of the feature itself~\cite{missmecha}.  
For continuous variables, extreme ranges (e.g., highest/lowest 10\%) receive elevated removal 
probability, simulating selective omission of clinically abnormal or rare values.  
For categorical variables, specific categories are assigned higher missingness rates; if the 
desired global rate is not met, additional categories are incrementally included.  
This setting captures self-dependent omission patterns that frequently arise in clinical and 
behavioral data.

By incorporating all three mechanisms at multiple missing rates, our evaluation protocol covers 
both random and systematically biased missingness, providing a rigorous and comprehensive 
benchmark for heterogeneous tabular imputation.

\begin{table}[t]
\caption{
Summary of benchmark datasets. Cate. and Cont. denote the number of categorical 
and continuous attributes, respectively. Classes indicates the number of target labels 
for downstream classification tasks, while -- denotes datasets used only for 
unsupervised evaluation.
}
\label{tab:dataset_summary}
\centering
\resizebox{0.48\textwidth}{!}{%
\begin{tabular}{rrrrr}
\toprule
\textbf{Dataset} & \textbf{Samples} & \textbf{Cate.} & \textbf{Cont.} & \textbf{Classes} \\
\midrule
Adult        & 48{,}842 & 7  & 7  & 2  \\ 
Banknote     & 1{,}372  & 0  & 4  & 2  \\
Breast       & 2{,}869  & 4  & 0  & 2  \\ 
Letter          & 20{,}000  & 0  & 16  & 26  \\

California         & 20{,}640 & 0 & 8 & Regression \\
E-commerce   & 10{,}999 & 7  & 3  & Regression \\ 
Student      & 649      & 11 & 2  & Regression  \\ 
Yacht        & 308      & 0  & 6  & Regression \\
\bottomrule
\end{tabular}
}
\end{table}

\subsection{Baselines and Evaluation Protocol}

To provide a comprehensive assessment of \texttt{MissHDM}, we compare it against three categories of representative imputation methods spanning statistical, deep generative, and diffusion-based paradigms.

\paragraph{Statistical Baselines.}
Classical imputers include \texttt{Mean/Mode} substitution, \texttt{MICE}\cite{mice}, and \texttt{MissForest}\cite{missforest}.
These approaches are computationally efficient and widely used in practice, but they model feature dependencies only through simple heuristics or shallow regression updates, making them an important baseline for evaluating the benefits of deep generative modeling.

\paragraph{Deep Generative Models.}
We include two representative latent-variable models: \texttt{GAIN}\cite{yoon2018gain}, a GAN-based imputer trained via adversarial learning, and \texttt{MIWAE}\cite{mattei2019miwae}, a VAE-based method optimized under an importance-weighted objective.
These methods capture nonlinear dependencies but typically assume homogeneous data types, requiring categorical variables to be discretized or one-hot encoded.

\paragraph{Diffusion-based Models.}
We evaluate three diffusion-based imputers: \texttt{CSDI}\cite{csdi}, its tabular variant \texttt{TabCSDI}\cite{tabcsdi}, and \texttt{MissDiff}~\cite{missdiff}.
All diffusion baselines follow standard stochastic DDPM sampling and use $T{=}100$ reverse steps with 100 stochastic samples per instance (median aggregation).
In contrast, \texttt{MissHDM} produces deterministic imputations in a single pass via its DDIM-based continuous channel and loopholed discrete channel.

\subsubsection{Imputation Error Metrics}

Following established practice in heterogeneous tabular imputation, we evaluate
performance using the \textbf{Average Imputation Error (AvgErr)}, defined as
\[
\text{AvgErr} = \frac{1}{D} \sum_{d=1}^{D} \mathrm{err}(d),
\]
where the per–feature error $\mathrm{err}(d)$ depends on the variable type.

\paragraph{Continuous variables}
For numerical features, we report the \textbf{ Root Mean Squared Error (RMSE)}:
\begin{equation}
\mathrm{err}(d)
= 
\sqrt{
\frac{1}{n_d}\sum_{i \in \mathcal{M}_d} (X_{id} - \hat{X}_{id})^{2}
}
\label{eq:nrmse}
\end{equation}

\paragraph{Categorical variables}
For discrete categorical features, reconstruction quality is measured by the
\textbf{accuracy error}:
\begin{equation}
\mathrm{err}(d)
=
\frac{1}{n_d}
\sum_{i \in \mathcal{M}_d}
\mathbb{I}(X_{id} \neq \hat{X}_{id}),
\label{eq:caterr}
\end{equation}
where $\hat{X}_{id} = \arg\max$ over the predicted categorical distribution.

\paragraph{Ordinal variables.}
If a categorical feature is ordinal with $R_d$ ordered levels, we additionally report the
\textbf{displacement error}:
\begin{equation}
\mathrm{err}(d)
=
\frac{1}{n_d}
\sum_{i \in \mathcal{M}_d}
\left|
\frac{X_{id} - \hat{X}_{id}}{R_d}
\right|.
\label{eq:orderr}
\end{equation}
This penalizes incorrect predictions proportionally to their ordinal distance.

\subsection{Implementation Details}

MissHDM is implemented using two diffusion backbones: one for continuous variables(Channel~1) and one for categorical/discrete variables (Channel~2).  Both channels operate on the same self-masked input and share the observed context $\mathbf{x}^{\mathrm{obs}}$ throughout training and sampling, ensuring that correlations across feature types are preserved.

\paragraph{Channel 1: Continuous diffusion}
The continuous channel adopts a feature-wise Transformer encoder based on the non-temporal
TabCSDI architecture~\citep{tabcsdi}.  
We retain only the feature-attention encoder and residual MLP blocks, which together
parameterize the noise-prediction network $\epsilon_\theta$.  
Inputs include the noisy continuous coordinates, a learned timestep embedding, and a mask
embedding indicating observed versus missing positions.  
This network supports deterministic DDIM sampling during inference.

\paragraph{Channel 2: Discrete loopholing diffusion}
For categorical and discrete variables, we use a lightweight feature-attention encoder that
operates on simplex-valued representations~\cite{loopholing}.
Each discrete value is encoded either as a one-hot vector or via an index embedding, which is mapped to a probability simplex through a softmax layer.   The encoder outputs the logits required for the reverse-time loopholing transition, ensuring that all intermediate states remain valid categorical distributions.

\paragraph{Joint Reconstruction}
Both channels receive identical conditioning signals consisting of observed feature values,
feature embeddings, and the binary mask.  
Conditioning is incorporated through cross-attention, allowing continuous and discrete channels
to utilize the same partial observations.  
During sampling, updated estimates from each channel are fed back to the shared conditioning module, enabling joint consistency between variable types.

\paragraph{Self-masking}
During training, we augment the original missing mask by randomly masking a subset of observed
entries to form pseudo-missing targets.  
The same self-masked input is supplied to both channels, enabling them to learn the conditional
distribution \( p(\mathbf{x}^{\mathrm{mis}} \mid \mathbf{x}^{\mathrm{obs}}) \) without relying on
external imputers.

\paragraph{Training settings}
Both channels are trained jointly using AdamW (learning rate $1\times 10^{-4}$, batch size 64,
weight decay $10^{-4}$).  
We use $T=100$ diffusion steps during training.  
Continuous noise levels follow a cosine $\beta_t$ schedule, while discrete transitions follow a
linearly annealed loopholing kernel.  
Training proceeds for 300 epochs with early stopping based on validation error.

\paragraph{Sampling}
At inference time, MissHDM uses 20-step deterministic DDIM updates for continuous variables
and 20-step reverse loopholing transitions for discrete variables.  
Final outputs are obtained by taking $\hat{\mathbf{x}}^{\mathrm{cont}}_0$ from the last DDIM
update and $\arg\max$ over discrete logits $\hat{\mathbf{p}}_{0}$.  
The reconstructed continuous and discrete variables are decoded and combined into the final
imputed sample.

\begin{table*}[!t]
\centering
\caption{
Regression performance (MAE $\downarrow$) under $30\%$, $50\%$, and $70\%$ missingness on four benchmark datasets. The best score for each setting is highlighted in $\best{bold}$, and the second-best result is marked with $\secondbest{underline}$.}
\label{tab:regression_results}
\resizebox{\textwidth}{!}{
\begin{tabular}{l|ccc|ccc|ccc|ccc|}
\hline
\textbf{Method} 
& \multicolumn{3}{c|}{\textbf{Yacht}} 
& \multicolumn{3}{c|}{\textbf{E-commerce}} 
& \multicolumn{3}{c|}{\textbf{California}} 
& \multicolumn{3}{c|}{\textbf{Student}} 
\\
\hline
\textbf{Rate} & 30\% & 50\% & 70\% & 30\% & 50\% & 70\% & 30\% & 50\% & 70\% & 30\% & 50\% & 70\% \\
\hline
Mean       & 1.1748& 1.3966& 1.5177
& 1.1264& 1.2585& 1.3947
& 0.5310& 0.6547& 0.7193
& 2.2286 & 2.3608 & 2.5107 \\
MICE       & 1.3032& 1.283& 1.5076
& 1.204& 1.218& 1.4373
& \secondbest{0.5009}& 0.6439& 0.7241
& 2.0147 & 2.1097 & 2.4936 \\
MissForest & 1.3681& 1.3306& 1.6776
& 1.2887& 1.2245& 1.5723
& 0.5098& \best{0.5832}& 0.7519
& 1.7569 & 2.0165 & 1.9655 \\
\hline
GAIN       & 1.0437& 1.2341& 1.3778
& 0.9787& 1.1108& 1.2736
& 0.5214& 0.713& 0.7384
& 1.2574 & 1.4025 & 1.5993 \\
MIWAE      & 1.046& 1.2262& 1.2757
& 0.9245& 1.1464& 1.1795
& 0.516& 0.7281& 0.7711
& 1.2347 & 1.4582 & 1.4614 \\\hline
CSDI     & 0.8049& 1.0274& 1.0349
& 0.7296& 0.9554& 0.8932
& 0.5191& 0.7085& \secondbest{0.7463}& 1.0257 & 1.2477 & 1.2098 \\

TabCSDI    & \secondbest{0.6397}& \best{0.9127}& 0.9578
& \best{0.5238}& \secondbest{0.8379}& \secondbest{0.8614}& 0.5103& \secondbest{0.6393}& 0.6908
& 0.8712 & 1.1371 & 1.1816 \\
MissDiff   & 0.7099& 0.9346& \secondbest{0.9477}& 0.5748& 0.8508& \best{0.8591}& 0.5242& 0.6565& 0.7481
& \secondbest{0.8571} & \secondbest{1.1297} & \best{1.1575} \\
\hline
MissHDD   & \best{0.5668}& \secondbest{0.9313}& \best{0.9203}& \secondbest{0.526}& \best{0.8054}& \secondbest{0.8641}& \best{0.4907}& 0.6422& \best{0.6827}& \best{0.8214} & \best{1.1143} & \secondbest{1.1713} \\
\hline
\end{tabular}
}
\end{table*}

\begin{table*}[!t]
\centering
\caption{
Classification performance (F1 Score $\uparrow$) under $30\%$, $50\%$, and $70\%$ missingness on four benchmark datasets. The best score for each setting is highlighted in $\best{bold}$, and the second-best result is marked with $\secondbest{underline}$.}
\label{tab:classification_results}
\resizebox{\textwidth}{!}{
\begin{tabular}{l|ccc|ccc|ccc|ccc|}
\hline
\textbf{Method} 
& \multicolumn{3}{c|}{\textbf{Adult}} 
& \multicolumn{3}{c|}{\textbf{Banknote}} 
& \multicolumn{3}{c}{\textbf{Breast}}& \multicolumn{3}{c|}{\textbf{Letter}} 
\\
\hline
\textbf{Rate} & 30\% & 50\% & 70\% & 30\% & 50\% & 70\% & 30\% & 50\% & 70\% & 30\% & 50\% & 70\% \\
\hline
Mean       & 0.4966 & 0.4750 & 0.4440 & 0.9000 & 0.7870 & 0.6703 & 0.7632& 0.6628& 0.5119
& 0.7960 & 0.6277 & 0.4046 \\
MICE       & 0.6734 & 0.6447 & 0.6218 & 0.9416 & 0.8414 & 0.7512 & 0.6906& 0.6301& 0.4932
& 0.7862 & 0.7850 & 0.4370 \\
MissForest & 0.6971 & 0.6517 & 0.6347 & 0.9517 & 0.8727 & 0.7821 & 0.7097& 0.6424& \secondbest{0.5243}& 0.8323 & 0.8148 & 0.4524 \\
\hline
GAIN       & 0.6048 & 0.5775 & 0.5441 & 0.9217 & 0.8725 & 0.7957 & 0.7178& 0.6399& 0.4961
& 0.8221 & 0.7567 & 0.4054 \\
MIWAE      & 0.6941 & 0.6327 & 0.6014 & 0.9226 & 0.8937 & 0.8125 & 0.7542& 0.5805& 0.5223
& 0.8674 & 0.8186 & 0.4253 \\\hline
CSDI     & 0.6827 & 0.6424 & 0.6218 & 0.9358 & 0.8867 & 0.8025 & 0.7305& 0.703& 0.5177
& 0.8479 & 0.8090 & 0.4249 \\

TabCSDI    & \secondbest{0.7214} & \secondbest{0.6851} & \best{0.6318} & \best{0.9527} & \secondbest{0.9014} & \secondbest{0.8657} & \secondbest{0.773}& 0.7278& \best{0.5282}& \secondbest{0.9049} & 0.8471 & 0.4467 \\
MissDiff   & 0.7046 & 0.6711 & 0.6247 & 0.9416 & 0.8867 & 0.8237 & 0.7445& \secondbest{0.7293}& 0.5185
& 0.8503 & 0.8242 & \secondbest{0.4541} \\
\hline
MissHDD   & \best{0.7228} & \best{0.6724} & \best{0.6318} & \best{0.9527} & \best{0.9214} & \best{0.8772} & \best{0.7873}& \best{0.7524}& 0.5202& \best{0.9117} & \best{0.8559} & \best{0.4581} \\
\hline
\end{tabular}
}
\end{table*}

\section{Results and Analysis}
\begin{figure*}[t]
    \centering
    
    \begin{subfigure}{0.32\linewidth}
        \centering
        \includegraphics[width=\linewidth, height=0.23\textheight]{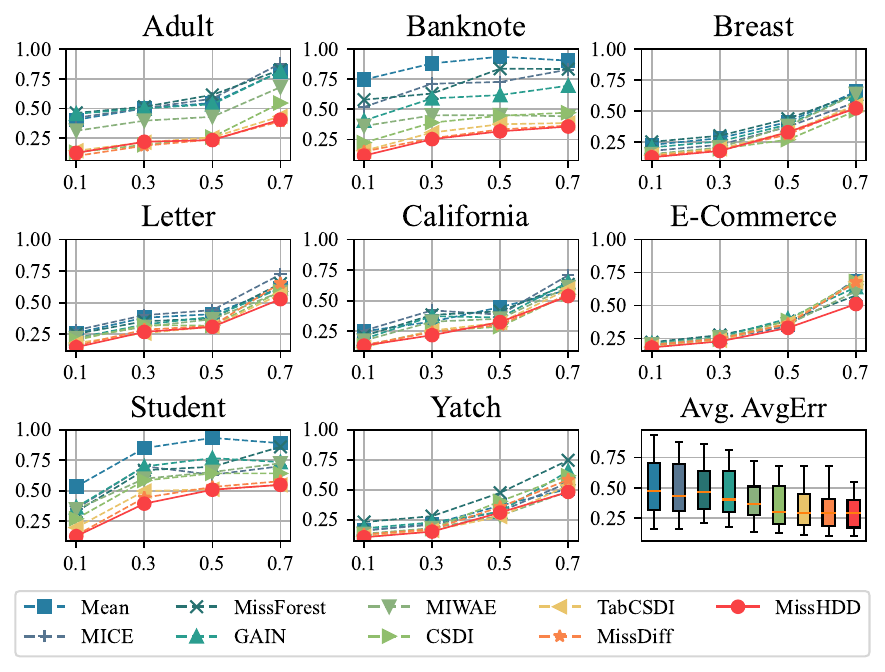}
        \caption{MCAR}
        \label{fig:rmse_mcar}
    \end{subfigure}
    \hfill
    \begin{subfigure}{0.32\linewidth}
        \centering
        \includegraphics[width=\linewidth, height=0.23\textheight]{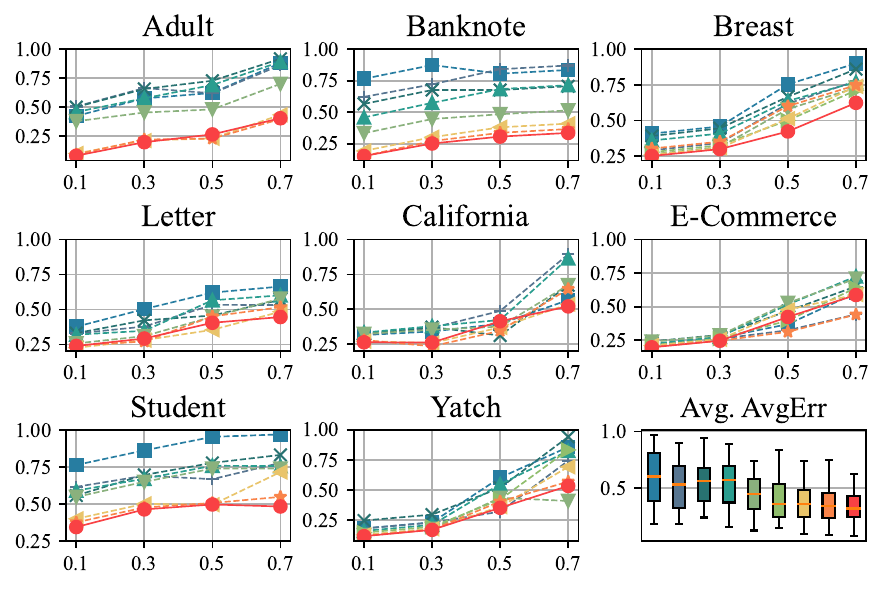}
        \caption{MAR}
        \label{fig:rmse_mar}
    \end{subfigure}
    \hfill
    \begin{subfigure}{0.32\linewidth}
        \centering
        \includegraphics[width=\linewidth, height=0.23\textheight]{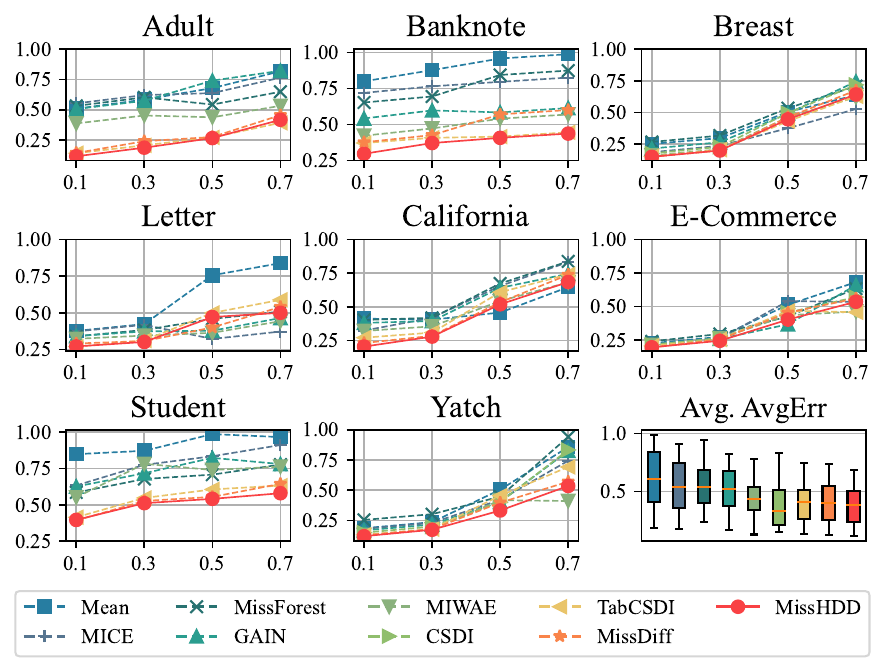}
        \caption{MNAR}
        \label{fig:rmse_mnar}
    \end{subfigure}

    \caption{
 Imputation accuracy comparison under three missingness mechanisms: \textbf{MCAR}, \textbf{MAR}, and \textbf{MNAR}.  
Each subplot visualizes dataset-wise AvgErr (left panel) and aggregated summary statistics (boxplot, right panel) across varying missing rates.
    }
    \label{fig:rmse_missingness_all}
\end{figure*}

\subsection{Imputation Accuracy}

We evaluate imputation quality across three missingness mechanisms—MCAR, MAR, and MNAR—each tested under four missing rates $(10\%, 30\%, 50\%, 70\%)$.
Figures~\ref{fig:rmse_mcar}, \ref{fig:rmse_mar}, and \ref{fig:rmse_mnar} report AvgErr across five benchmark datasets, together with aggregated boxplots summarising all settings.

Overall, the AvgErr of all baselines increases as the amount of missing data grows, especially under structurally dependent mechanisms (MAR and MNAR). Statistical imputers (\texttt{Mean}, \texttt{MICE}, \texttt{MissForest}) degrade quickly at higher missing rates, while deep generative methods (\texttt{GAIN}, \texttt{MIWAE}) show inconsistent performance across datasets. Existing diffusion-based approaches (\texttt{CSDI}, \texttt{TabCSDI}, \texttt{MissDiff}) are more stable but remain sensitive to discrete attributes and severe missingness.

In contrast, \texttt{MissHDD} maintains consistently low AvgErr across all mechanisms and missing rates. The deterministic continuous channel provides stable numerical reconstruction, and the loopholing-based discrete channel prevents the large categorical errors commonly seen in other diffusion models. The aggregated boxplots further show that \texttt{MissHDD} achieves both lower median AvgErr and substantially smaller variance, indicating stronger robustness to missingness severity and structure.

These findings demonstrate that heterogeneous diffusion with shared conditioning enables \texttt{MissHDD} to recover mixed-type tabular data reliably even when up to $70\%$ of the entries are absent.

\subsection{Downstream task utility}

We further evaluate the practical usefulness of the imputed data by measuring downstream predictive performance.
Table~\ref{tab:regression_results} reports MAE on four regression datasets (Yacht, E-commerce, California, Student) under $30\%$, $50\%$, $70\%$ missingness, while Table~\ref{tab:classification_results} summarises F1 scores on four classification benchmarks (Adult, Banknote, Breast, Letter).

Across both regression and classification tasks, MissHDD consistently matches or surpasses all competing methods, often achieving the best or second-best results in each setting. Conventional statistical imputers degrade substantially as missingness increases, and existing diffusion-based approaches exhibit notable sensitivity to discrete attributes or high missing rates. In contrast, MissHDD maintains stable and competitive performance even at $70\%$ missingness, reflecting the benefits of (i) deterministic numerical reconstruction and (ii) discrete-aware denoising that preserves categorical structure. The improvements in downstream F1 and MAE demonstrate that MissHDD not only imputes values more accurately but also retains task-relevant signal for subsequent predictive modelling—an essential criterion for real-world tabular analysis under severe missingness.

\subsection{Sampling Time}
Diffusion-based imputers such as CSDI, TabCSDI and MissDiff typically rely on
\emph{stochastic} DDPM sampling. In practice this means that, for each
incomplete instance, they generate many samples (we follow the common setting
of 100 samples) and aggregate them, e.g. by taking the median, in order to
obtain stable imputations. This strategy greatly increases inference cost.

In contrast, the diffusion channels in \texttt{MissHDD} use a
\emph{deterministic} DDIM-style sampler: a single trajectory is fully
specified by the initial noise and the learned score network. As a result,
one forward pass is sufficient and repeating the sampling process does not
change the output. Moreover, the number of diffusion steps $T$ can be reduced
without retraining, providing a direct knob to trade off accuracy and speed.

Table~\ref{tab:inference_summary} summarises average wall-clock time and
imputation RMSE across all benchmark datasets under a fixed hardware and
batch-size configuration. The first block reports DDPM-based baselines using
their standard 100-sample setting. The second block reports \texttt{MissHDD}
with a \emph{single} deterministic sample and different values of $T$.
With $T{=}100$, \texttt{MissHDD} is already slightly faster than the
stochastic baselines while achieving lower RMSE. Reducing the number of
steps to $T{=}50$ and $T{=}20$ yields $\sim$2$\times$ and $\sim$5$\times$
speed-ups respectively, with only a modest increase in error. These results
confirm that the deterministic design of \texttt{MissHDD} offers a favourable
accuracy–efficiency trade-off, making it particularly attractive for
large-scale or time-critical imputation scenarios.

\begin{table}[!t]
\centering
\caption{
Inference time and imputation accuracy (RMSE~$\downarrow$) for generative
imputation models. Baseline methods use 100 stochastic samples per instance,
while \texttt{MissHDD} uses a single deterministic DDIM trajectory with
different numbers of diffusion steps~$T$.
}
\label{tab:inference_summary}
\begin{tabular}{c|l|c|c}
\hline
\textbf{\#Samples} & \textbf{Method} 
& \textbf{Time (s) $\downarrow$}
& \textbf{RMSE $\downarrow$} \\
\hline
\multirow{3}{*}{100} 
& CSDI      & 820.73 & 0.3730 \\
& TabCSDI   & 794.21 & 0.3319 \\
& MissDiff  & 765.92 & 0.3163 \\
\hline
\multirow{3}{*}{1} 
& MissHDD ($T{=}100$) & 775.31 & \textbf{0.3051} \\
& MissHDD ($T{=}50$)  & 385.12 & 0.3167 \\
& MissHDD ($T{=}20$)  & \textbf{154.61} & 0.3343 \\
\hline
\end{tabular}
\end{table}

\begin{figure}[!h]
\centering
\begin{subfigure}[b]{0.325\columnwidth}
    \centering
    \includegraphics[width=\linewidth]{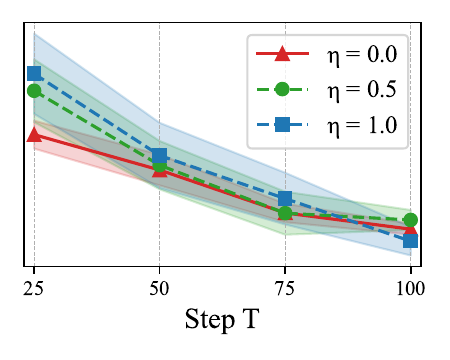}
    \vspace{-2em}
    \caption{Adult}
    \label{fig:ablation_sampling_adult}
\end{subfigure}
\hspace{-0.5em}
\begin{subfigure}[b]{0.325\columnwidth}
    \centering
    \includegraphics[width=\linewidth]{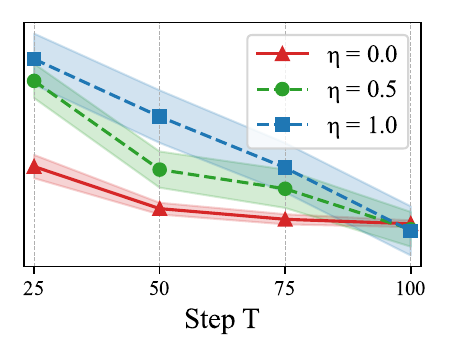}
            \vspace{-2em}
    \caption{Banknote}
    \label{fig:ablation_sampling_bank}
\end{subfigure}
\hspace{-0.5em}
\begin{subfigure}[b]{0.325\columnwidth}
    \centering
    \includegraphics[width=\linewidth]{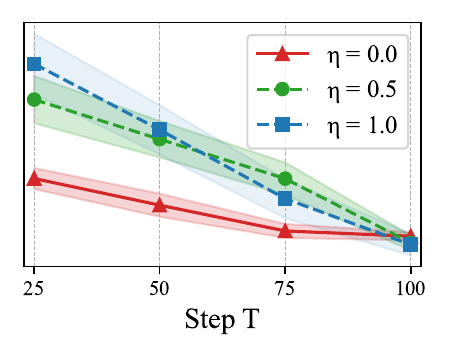}
            \vspace{-2em}
    \caption{California}
    \label{fig:ablation_sampling_cali}
\end{subfigure}
\begin{subfigure}[b]{0.325\columnwidth}
    \centering
    \includegraphics[width=\linewidth]{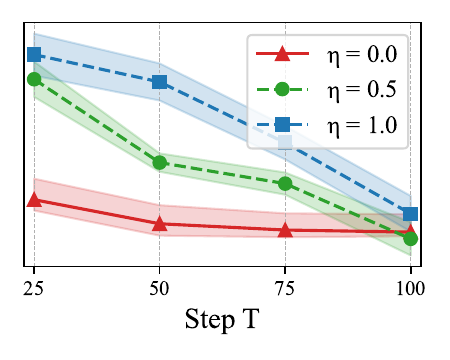}
    \caption{Letter}
    \label{fig:ablation_sampling_letter}
\end{subfigure}
\begin{subfigure}[b]{0.325\columnwidth}
    \centering
    \includegraphics[width=\linewidth]{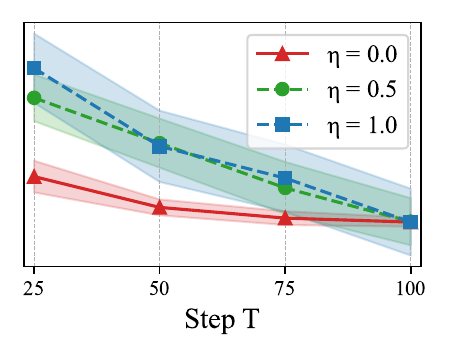}
    \caption{Student}
    \label{fig:ablation_sampling_student}
\end{subfigure}
\begin{subfigure}[b]{0.325\columnwidth}
    \centering
    \includegraphics[width=\linewidth]{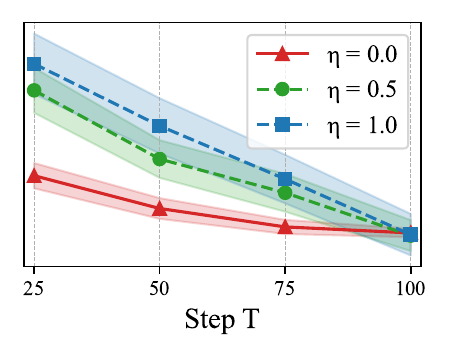}
    \caption{Average}
    \label{fig:ablation_sampling_avg}
\end{subfigure}
\caption{RMSE Comparison with varying $\eta$ values.}
\label{fig:ablation_sampling_combined}
\end{figure}

\section{Ablation Study}

\subsubsection{Synthetic heterogeneous dataset for ablation}

For this ablation study, we construct a synthetic
heterogeneous tabular dataset where both the data-generating process and the
missingness mechanisms are fully controlled.

We generate $n = 10{,}000$ samples with $d = 30$ features, split into
$d_{\mathrm{cont}} = 15$ continuous variables and $d_{\mathrm{cat}} = 15$
categorical variables. Let $\mathbf{z} \in \mathbb{R}^{15}$ denote a latent
Gaussian vector sampled as
\[
\mathbf{z} \sim \mathcal{N}(\mathbf{0}, \mathbf{I}_{15}).
\]

\paragraph{Continuous features.}
Each continuous feature is generated as a nonlinear transformation of
$\mathbf{z}$ with added Gaussian noise:
\[
x^{\mathrm{cont}}_k
= \mathbf{a}_k^\top \mathbf{z}
  + b_k \,\phi(\mathbf{c}_k^\top \mathbf{z})
  + \varepsilon_k,
\qquad
\varepsilon_k \sim \mathcal{N}(0, \sigma^2),
\]
for $k = 1,\dots,d_{\mathrm{cont}}$, where
$\phi(\cdot)$ is a smooth nonlinearity (e.g., $\tanh$),
and $\mathbf{a}_k, \mathbf{c}_k \in \mathbb{R}^4$, $b_k \in \mathbb{R}$ and
$\sigma > 0$ are fixed coefficients. This construction induces correlated,
non-Gaussian continuous features, mimicking realistic tabular structure.

\paragraph{Categorical features.}
Each categorical feature $x^{\mathrm{cat}}_j$ takes values in
$\{1,\dots,K_j\}$ with $K_j \in \{3,4,5\}$.
We first compute class logits as linear functions of the shared latent vector:
\[
\boldsymbol{\ell}_j = W_j \mathbf{z} + \boldsymbol{\xi}_j,
\qquad
\boldsymbol{\xi}_j \sim \mathcal{N}(\mathbf{0}, \sigma_{\text{cat}}^2 \mathbf{I}),
\]
where $W_j \in \mathbb{R}^{K_j \times 4}$.
Class probabilities are obtained via a softmax,
$\boldsymbol{\pi}_j = \mathrm{softmax}(\boldsymbol{\ell}_j)$, and the
categorical outcome is sampled as
\[
x^{\mathrm{cat}}_j \sim \mathrm{Cat}(\boldsymbol{\pi}_j),
\qquad j = 1,\dots,d_{\mathrm{cat}}.
\]
This yields categorical variables that are strongly but nonlinearly
correlated with the continuous ones through the shared latent vector
$\mathbf{z}$.

\paragraph{Missingness mechanisms}
Starting from the fully observed synthetic dataset $\mathbf{X}$, we generate
missingness masks $\mathbf{M}$ under the three canonical mechanisms
(MCAR, MAR, MNAR) at target missing rates
$\rho \in \{0.3, 0.5\}$:
\begin{itemize}
    \item \textbf{MCAR:} each entry is independently set to missing with
    probability $\rho$, independent of all feature values.
    \item \textbf{MAR:} for each feature $v$, the missingness probability
    depends on another observed feature $u$
    (e.g., larger values of $u$ induce a higher missing probability for $v$),
    following the threshold-based construction used in our real-data
    experiments.
    \item \textbf{MNAR:} missingness depends on the value of the feature itself:
    for continuous variables, extreme values (e.g., top/bottom quantiles) are
    removed with higher probability; for categorical variables, a subset of
    categories is assigned elevated missingness probability until the target
    rate $\rho$ is reached.
\end{itemize}

\subsection{Contribution of the Two-Channel Architecture}

\begin{table}[!t]
\centering
\caption{Ablation on the two-channel architecture under MCAR with 30\% and 70\% missingness.
\textbf{Full} denotes the complete MissHDD model; 
\textbf{Cont.} is a continuous-only DDIM imputer applied to numerical features only; 
\textbf{Disc.} is a discrete-only imputer that discretises continuous features via binning.
The evaluation metric is \textbf{AvgErr}:  
for continuous-only settings it reduces to RMSE,  
and for categorical-only settings it reduces to classification error.  For mixed-type settings, AvgErr is computed by averaging RMSE over numerical
dimensions and cross-entropy error over categorical dimensions.}

\label{tab:ablation_two_channel}
\resizebox{0.80\textwidth}{!}{%
\begin{tabular}{l l c c c}
\toprule
Dataset & Model 
& MCAR-AvgErr $\downarrow$ 
& MAR-AvgErr $\downarrow$ 
& MNAR-AvgErr $\downarrow$ \\
\midrule
\multirow{3}{*}{Adult (30\%)} 
& Cont.-only  & 0.2345 & 0.2323 & 0.2523 \\
& Disc.-only  & 0.2574 & 0.3214 & 0.3124 \\
& \textbf{Full} & \best{0.2202} & \best{0.1956} & \best{0.1864} \\
\midrule
\multirow{3}{*}{Adult (70\%)} 
& Cont.-only  & 0.4312 & 0.4314 & 0.4552 \\
& Disc.-only  & 0.4712 & 0.5027 & 0.5632 \\
& \textbf{Full} & \best{0.4063} & \best{0.4029} & \best{0.4167} \\
\midrule
\multirow{3}{*}{Synthetic (30\%)} 
& Cont.-only  & 0.3837 & 0.3758 & 0.4248 \\
& Disc.-only  & 0.4521 & 0.4693 & 0.4851 \\
& \textbf{Full} & \best{0.3459} & \best{0.3581} & \best{0.3948} \\
\midrule
\multirow{3}{*}{Synthetic (70\%)} 
& Cont.-only  & 0.6421 & 0.7047 & 0.7623 \\
& Disc.-only  & 0.6821 & 0.7948 & 0.8124 \\
& \textbf{Full} & \best{0.5821} & \best{0.6471} & \best{0.7123} \\
\bottomrule
\end{tabular}}
\end{table}

To quantify the benefit of modelling continuous and categorical variables through distinct diffusion pathways, we compare \texttt{MissHDD} with two single-channel variants: (i) a continuous-only model where all features are numerically encoded and imputed via a DDIM-based diffusion process; and (ii) a discrete-only model where continuous features are discretised into quantile bins and imputed using the discrete diffusion channel.

Across all datasets, the full two-channel architecture consistently achieves lower imputation error and higher categorical accuracy. The continuous-only variant performs poorly on categorical variables due to semantic drift induced by continuous relaxations, whereas the discrete-only variant loses precision on numerical attributes due to discretisation artefacts. These results confirm that heterogeneous tabular data require separate generative mechanisms and validate the design choice underlying \texttt{MissHDD}.

\subsection{Self-mask ratio and effectiveness}

\begin{table}[!t]
\centering
\caption{
Ablation of the self-masking strategy on a synthetic dataset 
(with 50\% MCAR missingness). 
\textbf{Zero-Imp} and \textbf{Mean-Imp} denote training without any additional
masking, where missing entries are filled using zero-imputation or mean-imputation, respectively. 
All remaining rows apply extra self-masking to a random 10\%, 30\%, or 50\% subset 
of observed entries under different masking schemes. 
This evaluates how exposing the model to diverse pseudo-missing patterns during training 
affects robustness and overall imputation accuracy.
}
\label{tab:ablation_self_mask}
\resizebox{0.8\textwidth}{!}{
\begin{tabular}{c l c c c}
\toprule
\textbf{Self-mask Strategy} 
& Ratio
& \textbf{MCAR-AvgEr $\downarrow$}
& \textbf{MAR-AvgEr $\downarrow$}
& \textbf{MNAR-AvgEr $\downarrow$} \\
\midrule

\multirow{2}{*}{\textbf{No Self-mask}} 
& Zero-Imp & 0.4231 & 0.4528 & 0.4612 \\
& Mean-Imp & 0.3545 & 0.3681 & 0.4048\\
\midrule

\multirow{3}{*}{\textbf{MCAR Self-mask}} 
& 10\%  & 0.3634 & 0.3728 & 0.4124 \\
& 30\%  & \best{0.3459} & 0.3581 & 0.3948 \\
& 50\%  & \best{0.3459} & \best{0.3501} & \best{0.3930} \\
\midrule

\multirow{3}{*}{\textbf{MAR Self-mask}} 
& 10\%  & 0.3759 & 0.3874 & 0.3852 \\
& 30\%  & \best{0.3532} & \best{0.3524} & \best{0.3749}\\
& 50\%  & 0.3592 & 0.3651 & 0.3824\\
\midrule

\multirow{3}{*}{\textbf{MNAR Self-mask}} 
& 10\%  & 0.3682 & 0.3839 & 0.4029\\
& 30\%  & \best{0.3524} & \best{0.3624} & 0.3849\\
& 50\%  & 0.3759 & 0.3749 & \best{0.3833}\\
\bottomrule
\end{tabular}}
\end{table}

Table \ref{tab:ablation_self_mask} presents a detailed ablation of our self-masking design.
We compare training without any auxiliary masking (“Zero-Imp” and “Mean-Imp”) against various self-masking ratios $(10\%, 30\%, 50\%)$ applied under MCAR, MAR, and MNAR masking schemes.

First, models trained without self-masking exhibit the weakest performance across all three missingness mechanisms. This confirms that relying solely on the original mask causes the model to overfit specific missingness patterns and generalise poorly to unseen masks—particularly evident under MAR and MNAR, where the distribution shift between train-time and test-time masks is substantial.

Introducing self-masking consistently improves AvgErr across all settings.
Among all ratios, $30\%$ self-masking yields the most stable and competitive performance, striking a balance between providing sufficient pseudo-missing supervision and avoiding excessive corruption of observed features.
Although $50\%$ masking can achieve comparable or slightly better results in some cases, we observe slower convergence and higher optimisation variance due to the reduced amount of intact conditioning information—an effect amplified on small synthetic datasets.
Conversely, $10\%$ masking is generally insufficient: the model sees too few pseudo-missing entries to learn robust conditional distributions.

Comparing different masking mechanisms (MCAR-, MAR-, MNAR-style self-masking), we observe no consistently dominant masking strategy across the evaluated settings. While MAR/MNAR masking occasionally offer small gains, the differences are marginal. For simplicity and robustness, MCAR-style self-masking emerges as the most reliable and computationally convenient choice, providing strong and consistent performance across all test mechanisms.

Overall, these results demonstrate that self-masking is an essential component of our training pipeline: it improves robustness, stabilises training, and yields consistently lower errors without requiring additional clean data.

\subsection{Effect of Sampling Stochasticity and Diffusion Steps}

\begin{table}[!t]
\centering
\caption{Impact of sampling stochasticity ($\eta$) and number of sampling steps ($T$) on RMSE under 50\% missing rate.
}
\label{tab:eta_step_table}
\begin{tabular}{c|cc|cc|cc}
\hline
\multirow{2}{*}{$\eta$} & \multicolumn{2}{c|}{\textbf{Letter}} & \multicolumn{2}{c|}{\textbf{California}} & \multicolumn{2}{c}{\textbf{Average}} \\
\cline{2-7}
 & 25  & 100 & 25  & 100 & 25  & 100 \\
\hline
0.0 & \textbf{0.3214}  & 0.2254 & \textbf{0.5014}  & 0.3254 &\textbf{0.4977} & 0.3293\\
0.5 & 0.6780  & \textbf{0.2054} & 0.7421  & 0.3054 &0.7412  & \textbf{0.3167}\\
1.0 & 0.7510  & 0.2796 & 0.8510  & \textbf{0.2996} &0.8170 & 0.3343\\
\hline
\end{tabular}
\end{table}

To understand the behavior of the continuous diffusion channel under different inference-time configurations, we examine how sampling stochasticity and the number of reverse steps influence imputation quality. Sampling stochasticity is controlled by the DDIM noise parameter $\eta \in \{0.0, 0.5, 1.0\}$, where $\eta{=}0$ yields a fully deterministic trajectory and larger values progressively approach standard DDPM stochastic sampling. The variance injected at each reverse step $\tau_i$ is given by:
\[
\sigma_{\tau_i}(\eta)
= \eta 
\sqrt{\frac{1 - \alpha_{\tau_{i-1}}}{1 - \alpha_{\tau_i}}}
\cdot
\sqrt{1 - \frac{\alpha_{\tau_i}}{\alpha_{\tau_{i-1}}}},
\]
which allows us to isolate the effect of stochasticity while keeping model parameters fixed.

We further vary the number of sampling steps 
$T \in \{25, 50, 75, 100\}$ 
to evaluate the trade-off between computational cost and reconstruction fidelity. This setting reflects practical deployment scenarios where latency constraints may limit the number of diffusion steps.

Table~\ref{tab:eta_step_table} and Figure~\ref{fig:ablation_sampling_combined} report the RMSE and its standard deviation across multiple runs. Consistently, the deterministic setting ($\eta{=}0$) produces near-zero variance and achieves competitive or superior accuracy, especially when $T$ is small. Introducing stochasticity ($\eta>0$) can marginally improve RMSE at larger $T$, but at the cost of substantially higher variance and longer inference time. These results highlight that the deterministic sampling strategy used in the continuous channel of \texttt{MissHDD} provides an effective balance between stability, accuracy, and efficiency.

\section{Conclusion}

We introduced \textbf{MissHDD}, a unified hybrid diffusion framework for imputing heterogeneous, incomplete tabular data. By coupling a deterministic DDIM-based continuous channel with a loopholing-based discrete diffusion channel, MissHDD models numerical and categorical/discrete variables on their native manifolds while maintaining coherent conditioning through a shared observed context. This two-channel design allows MissHDD to handle mixed-type data without continuous relaxations and to produce stable, high-fidelity imputations in a single deterministic pass.

Through extensive experiments across multiple missingness mechanisms (MCAR, MAR, MNAR), missing rates, and downstream predictive tasks, we demonstrate that MissHDD consistently achieves state-of-the-art accuracy, strong downstream utility, and markedly improved output stability relative to stochastic diffusion models. Its deterministic sampling also yields substantial reductions in inference time, making the method well-suited for real-time, large-scale, or deployment-oriented settings.

Future work includes extending MissHDD to handle richer variable families (e.g., ordinal, multi-valued, or time-aware features), developing tighter correlations between diffusion channels, and exploring adaptive masking strategies that better match real-world missingness behaviour. Overall, MissHDD provides an effective, principled, and practical solution for heterogeneous data imputation, opening new directions for diffusion models beyond conventional continuous domains.

\bibliographystyle{elsarticle-num}
\bibliography{mybib}

@article{Deeplearningisnotallyouneed,
  title={Tabular data: Deep learning is not all you need},
  author={Shwartz-Ziv, Ravid and Armon, Amitai},
  journal={Information Fusion},
  volume={81},
  pages={84--90},
  year={2022},
  publisher={Elsevier}
}

@inproceedings{diffputer,
title={DiffPuter: An {EM}-Driven Diffusion Model for Missing Data Imputation},
author={Hengrui Zhang and Liancheng Fang and Qitian Wu and Philip S. Yu},
booktitle={The Thirteenth International Conference on Learning Representations},
year={2025},
url={https://openreview.net/forum?id=3fl1SENSYO}
}

@inproceedings{dm1,
  title={Diffir: Efficient diffusion model for image restoration},
  author={Xia, Bin and Zhang, Yulun and Wang, Shiyin and Wang, Yitong and Wu, Xinglong and Tian, Yapeng and Yang, Wenming and Van Gool, Luc},
  booktitle={Proceedings of the IEEE/CVF International Conference on Computer Vision},
  pages={13095--13105},
  year={2023}
}

@inproceedings{dm2,
  title={Generative diffusion prior for unified image restoration and enhancement},
  author={Fei, Ben and Lyu, Zhaoyang and Pan, Liang and Zhang, Junzhe and Yang, Weidong and Luo, Tianyue and Zhang, Bo and Dai, Bo},
  publisher={Proceedings of the IEEE/CVF conference on computer vision and pattern recognition},
  pages={9935--9946},
  year={2023}
}

@article{dreview1,
  title={Diffusion models in vision: A survey},
  author={Croitoru, Florinel-Alin and Hondru, Vlad and Ionescu, Radu Tudor and Shah, Mubarak},
  journal={IEEE Transactions on Pattern Analysis and Machine Intelligence},
  volume={45},
  number={9},
  pages={10850--10869},
  year={2023},
  publisher={IEEE}
}

@InProceedings{MLDL,
author="Zhou, Youran
and Bouadjenek, Mohamed Reda
and Aryal, Sunil",
editor="Bifet, Albert
and Krilavi{\v{c}}ius, Tomas
and Miliou, Ioanna
and Nowaczyk, Slawomir",
title="Missing Data Imputation: Do Advanced ML/DL Techniques Outperform Traditional Approaches?",
booktitle="Machine Learning and Knowledge Discovery in Databases. Applied Data Science Track",
year="2024",
publisher="Springer Nature Switzerland",
address="Cham",
pages="100--115",
isbn="978-3-031-70381-2"
}

@inproceedings{ddpm,
author = {Ho, Jonathan and Jain, Ajay and Abbeel, Pieter},
title = {Denoising diffusion probabilistic models},
year = {2020},
isbn = {9781713829546},
publisher = {Curran Associates Inc.},
address = {Red Hook, NY, USA},
booktitle = {Proceedings of the 34th International Conference on Neural Information Processing Systems},
articleno = {574},
numpages = {12},
location = {Vancouver, BC, Canada},
series = {NIPS '20}
}

@article{knn,
  title={Nearest neighbor selection for iteratively kNN imputation},
  author={Zhang, Shichao},
  journal={Journal of Systems and Software},
  volume={85},
  number={11},
  pages={2541--2552},
  year={2012},
  publisher={Elsevier}
}

@article{misGAN,
  title={Improved generative adversarial network with deep metric learning for missing data imputation},
  author={Al-taezi, Mohammed Ali and Wang, Yu and Zhu, Pengfei and Hu, Qinghua and Al-Badwi, Abdulrahman},
  journal={Neurocomputing},
  volume={570},
  pages={127062},
  year={2024},
  publisher={Elsevier}
}

@inproceedings{mean,
  title={Handling missing data problems with sampling methods},
  author={Houari, Rima and Bounceur, Ahc{\`e}ne and Tari, A Kamel and Kecha, M Tahar},
  publisher={2014 International conference on advanced networking distributed systems and applications},
  pages={99--104},
  year={2014},
  organization={IEEE}
}

@article{mean2,
  title={Missing data imputation techniques},
  author={Song, Qinbao and Shepperd, Martin},
  journal={International journal of business intelligence and data mining},
  volume={2},
  number={3},
  pages={261--291},
  year={2007},
  publisher={Inderscience Publishers}
}

@article{grape,
  title={Handling missing data with graph representation learning},
  author={You, Jiaxuan and Ma, Xiaobai and Ding, Yi and Kochenderfer, Mykel J and Leskovec, Jure},
  journal={Advances in Neural Information Processing Systems},
  volume={33},
  pages={19075--19087},
  year={2020}
}

@inproceedings{igrm,
  title={Data imputation with iterative graph reconstruction},
  author={Zhong, Jiajun and Gui, Ning and Ye, Weiwei},
  booktitle={Proceedings of the AAAI Conference on Artificial Intelligence},
  volume={37},
  number={9},
  pages={11399--11407},
  year={2023}
}

@article{missforest,
   title={MissForest—non-parametric missing value imputation for mixed-type data},
   volume={28},
   ISSN={1367-4803},
   url={http://dx.doi.org/10.1093/bioinformatics/btr597},
   DOI={10.1093/bioinformatics/btr597},
   number={1},
   journal={Bioinformatics},
   publisher={Oxford University Press (OUP)},
   author={Stekhoven, Daniel J. and Bühlmann, Peter},
   year={2011},
   month=oct, pages={112–118} }

@inproceedings{csdi,
author = {Tashiro, Yusuke and Song, Jiaming and Song, Yang and Ermon, Stefano},
title = {CSDI: conditional score-based diffusion models for probabilistic time series imputation},
year = {2021},
isbn = {9781713845393},
publisher = {Curran Associates Inc.},
address = {Red Hook, NY, USA},
booktitle = {Proceedings of the 35th International Conference on Neural Information Processing Systems},
articleno = {1900},
numpages = {13},
series = {NIPS '21}
}

@article{sensor,
  title={Missing data problem in the monitoring system: A review},
  author={Du, Jinghan and Hu, Minghua and Zhang, Weining},
  journal={IEEE Sensors Journal},
  volume={20},
  number={23},
  pages={13984--13998},
  year={2020},
  publisher={IEEE}
}

@article{recommendation,
  title={Recommendation system based on deep learning methods: a systematic review and new directions},
  author={Da’u, Aminu and Salim, Naomie},
  journal={Artificial Intelligence Review},
  volume={53},
  number={4},
  pages={2709--2748},
  year={2020},
  publisher={Springer}
}

@article{financial,
  title={Missing financial data},
  author={Bryzgalova, Svetlana and Lerner, Sven and Lettau, Martin and Pelger, Markus},
  journal={The Review of Financial Studies},
  volume={38},
  number={3},
  pages={803--882},
  year={2025},
  publisher={Oxford University Press}
}

@article{healthcare,
  title={Data completeness in healthcare: a literature survey},
  author={Liu, Caihua and Talaei-Khoei, Amir and Zowghi, Didar and Daniel, Jay},
  journal={Pacific Asia Journal of the Association for Information Systems},
  volume={9},
  number={2},
  pages={5},
  year={2017}
}

@inproceedings{
ddim,
title={Denoising Diffusion Implicit Models},
author={Jiaming Song and Chenlin Meng and Stefano Ermon},
booktitle={International Conference on Learning Representations},
year={2021},
url={https://openreview.net/forum?id=St1giarCHLP}
}

@inproceedings{missddim,
author = {Zhou, Youran and Bouadjenek, Mohamed Reda and Aryal, Sunil},
title = {MissDDIM: Deterministic and Efficient Conditional Diffusion for Tabular Data Imputation},
year = {2025},
isbn = {9798400720406},
publisher = {Association for Computing Machinery},
address = {New York, NY, USA},
url = {https://doi.org/10.1145/3746252.3760943},
doi = {10.1145/3746252.3760943},
booktitle = {Proceedings of the 34th ACM International Conference on Information and Knowledge Management},
pages = {5525–5529},
numpages = {5},
location = {Seoul, Republic of Korea},
series = {CIKM '25}
}

@article{campbell2022continuous,
  title={A continuous time framework for discrete denoising models},
  author={Campbell, Andrew and Benton, Joe and De Bortoli, Valentin and Rainforth, Thomas and Deligiannidis, George and Doucet, Arnaud},
  journal={Advances in Neural Information Processing Systems},
  volume={35},
  pages={28266--28279},
  year={2022}
}

@misc{missmecha,
      title={MissMecha: An All-in-One Python Package for Studying Missing Data Mechanisms}, 
      author={Youran Zhou and Mohamed Reda Bouadjenek and Sunil Aryal},
      year={2025},
      eprint={2508.04740},
      archivePrefix={arXiv},
      primaryClass={cs.LG},
      url={https://arxiv.org/abs/2508.04740}, 
}

@misc{loopholing,

      title={Loopholing Discrete Diffusion: Deterministic Bypass of the Sampling Wall}, 

      author={Mingyu Jo and Jaesik Yoon and Justin Deschenaux and Caglar Gulcehre and Sungjin Ahn},

      year={2025},

      eprint={2510.19304},

      archivePrefix={arXiv},

      primaryClass={cs.LG},

      url={https://arxiv.org/abs/2510.19304}, 

}

@inproceedings{sun2022score,
  title={Score-based Continuous-time Discrete Diffusion Models},
  author={Sun, Haoran and Yu, Lijun and Dai, Bo and Schuurmans, Dale and Dai, Hanjun},
  booktitle={The Eleventh International Conference on Learning Representations},
  year={2022}
}

@inproceedings{tabddpm,
author = {Kotelnikov, Akim and Baranchuk, Dmitry and Rubachev, Ivan and Babenko, Artem},
title = {TabDDPM: modelling tabular data with diffusion models},
year = {2023},
publisher = {JMLR.org},
booktitle = {Proceedings of the 40th International Conference on Machine Learning},
articleno = {725},
numpages = {16},
location = {Honolulu, Hawaii, USA},
series = {ICML'23}
}

@article{mice,
  title={Men and mice: relating their ages},
  author={Dutta, Sulagna and Sengupta, Pallav},
  journal={Life sciences},
  volume={152},
  pages={244--248},
  year={2016},
  publisher={Elsevier}
}

@article{hivae,
  title={Handling incomplete heterogeneous data using vaes},
  author={Nazabal, Alfredo and Olmos, Pablo M and Ghahramani, Zoubin and Valera, Isabel},
  journal={Pattern Recognition},
  volume={107},
  pages={107501},
  year={2020},
  publisher={Elsevier}
}

@inproceedings{mattei2019miwae,
  title={MIWAE: Deep generative modelling and imputation of incomplete data sets},
  author={Mattei, Pierre-Alexandre and Frellsen, Jes},
  booktitle={International conference on machine learning},
  pages={4413--4423},
  year={2019},
  organization={PMLR}
}

@inproceedings{yoon2018gain,
  title={Gain: Missing data imputation using generative adversarial nets},
  author={Yoon, Jinsung and Jordon, James and Schaar, Mihaela},
  booktitle={International conference on machine learning},
  pages={5689--5698},
  year={2018},
  organization={PMLR}
}

@inproceedings{ot,
  title={Missing data imputation using optimal transport},
  author={Muzellec, Boris and Josse, Julie and Boyer, Claire and Cuturi, Marco},
  booktitle={International Conference on Machine Learning},
  pages={7130--7140},
  year={2020},
  organization={PMLR}
}

@article{MNAR,
  title={Inference and missing data},
  author={Rubin, Donald B},
  journal={Biometrika},
  volume={63},
  number={3},
  pages={581--592},
  year={1976},
  publisher={Oxford University Press}
}

@article{tablegan,
   title={Data synthesis based on generative adversarial networks},
   volume={11},
   ISSN={2150-8097},
   url={http://dx.doi.org/10.14778/3231751.3231757},
   DOI={10.14778/3231751.3231757},
   number={10},
   journal={Proceedings of the VLDB Endowment},
   publisher={Association for Computing Machinery (ACM)},
   author={Park, Noseong and Mohammadi, Mahmoud and Gorde, Kshitij and Jajodia, Sushil and Park, Hongkyu and Kim, Youngmin},
   year={2018},
   month=jun, pages={1071–1083} }

@inproceedings{ctgan,
 author = {Xu, Lei and Skoularidou, Maria and Cuesta-Infante, Alfredo and Veeramachaneni, Kalyan},
 booktitle = {Advances in Neural Information Processing Systems},
 editor = {H. Wallach and H. Larochelle and A. Beygelzimer and F. d\textquotesingle Alch\'{e}-Buc and E. Fox and R. Garnett},
 pages = {},
 publisher = {Curran Associates, Inc.},
 title = {Modeling Tabular data using Conditional GAN},
 url = {https://proceedings.neurips.cc/paper_files/paper/2019/file/254ed7d2de3b23ab10936522dd547b78-Paper.pdf},
 volume = {32},
 year = {2019}
}

@InProceedings{tvae,
author="Kataria, Vipin
and Kumar, Nitin
and Patel, Parth",
editor="Zhang, Shunli
and Zhang, Yu
and Wibowo, Santoso
and Zhang, Liang-Jie",
title="Environmental Data Imputation via Temporal VAE with Learned Missing Value Representations",
booktitle="Big Data -- BigData 2025",
year="2025",
publisher="Springer Nature Switzerland",
address="Cham",
pages="207--220",
isbn="978-3-032-06524-7"
}

@misc{missdiff,
      title={MissDiff: Training Diffusion Models on Tabular Data with Missing Values}, 
      author={Yidong Ouyang and Liyan Xie and Chongxuan Li and Guang Cheng},
      year={2023},
      eprint={2307.00467},
      archivePrefix={arXiv},
      primaryClass={cs.LG},
      url={https://arxiv.org/abs/2307.00467}, 
}

@inproceedings{tabcsdi,
  title={Diffusion models for missing value imputation in tabular data},
  author={Zheng, Shuhan and Charoenphakdee, Nontawat},
  booktitle={NeurIPS Table Representation Learning (TRL) Workshop},
  year={2022}
}
\end{document}